%% file: main.tex
\documentclass[runningheads]{llncs}

 \usepackage{eccv_supp}

\usepackage{eccvabbrv}

\usepackage{graphicx}
\usepackage{wrapfig}
\usepackage{booktabs}
\usepackage{algorithm}
\usepackage{algpseudocode}
\usepackage{enumitem}
\usepackage[table]{xcolor}
\usepackage[skins,breakable]{tcolorbox}
\usepackage{multirow}
\usepackage{multicol}
\usepackage{fontawesome5}
\usepackage{xcolor}
\usepackage{bm}

\DeclareMathOperator*{\argmax}{argmax}
\definecolor{LightGray}{RGB}{240,240,240}
\definecolor{LightBlue}{RGB}{246,249,253}
\definecolor{myblue}{RGB}{102, 138, 174}
\newcommand{\inlineicon}[3]{%
  \raisebox{#3}{\includegraphics[height=#2]{#1}}%
}
\newtcolorbox{mybox}[1]{colback=myblue!10!white,colframe=myblue,fonttitle=\bfseries,title=#1}
\def\CircleArrowright{\ensuremath{%
  \rotatebox[origin=c]{310}{$\circlearrowright$}}}
\newcommand{\vlnbert}{VLN$\protect\CircleArrowright$BERT}
\usepackage[accsupp]{axessibility}  

\usepackage[pagebackref,breaklinks,colorlinks,citecolor=eccvblue]{hyperref}

\usepackage{orcidlink}

\begin{document}

\title{\inlineicon{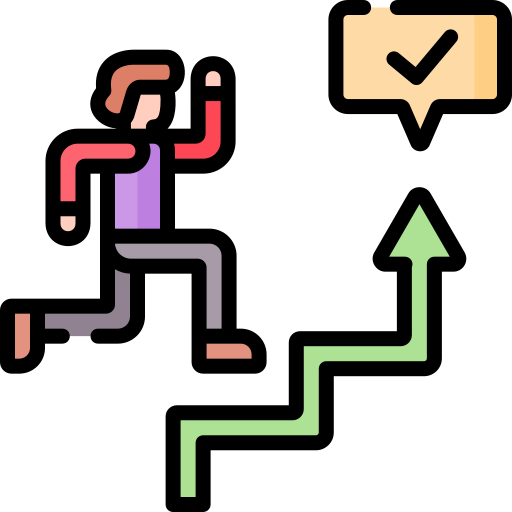}{1.3em}{-0.5em}Let's Reward Step-by-Step: Step-Aware Contrastive Alignment for Vision-Language Navigation in Continuous Environments} 

\titlerunning{\inlineicon{pictures/career.png}{1.3em}{-0.5em}Let's Reward Step-by-Step}

\author{Haoyuan Li\inst{1} \and
Rui Liu\inst{1} \and
Hehe Fan\inst{1} \and 
Yi Yang\inst{1}${\dagger}$}

\authorrunning{H. Li et al.}

\institute{College of Computer Science and Technology, Zhejiang University, Hangzhou, China \email{(haoyuanli, ruiliu, hehefan, yangyics)@zju.edu.cn} \\
\url{https://github.com/lhy-zjut/SACA}}

\maketitle

\input{sections/Abstract}
\input{sections/Introduction}
\input{sections/Related_work}
\input{sections/Method}
\input{sections/Experiments}

\input{sections/Conclusion}


%
%

\bibliographystyle{splncs04}
\bibliography{main}

\appendix
\input{sections/Appendix}
\end{document}

%% file: sections/Abstract.tex
\begin{abstract}
Vision-Language Navigation in Continuous Environments 

(VLN-CE) requires agents to learn complex reasoning from long-horizon human interactions. While Multi-modal Large Language Models (MLLMs) have driven recent progress, current training paradigms struggle to balance generalization capability, error recovery and training stability. Specifically, \textbf{(i)} policies derived from SFT suffer from compounding errors, struggling to recover from out-of-distribution states, and  \textbf{(ii)} Reinforcement Fine-Tuning (RFT) methods \textit{e.g.} GRPO are bottlenecked by sparse outcome rewards. Their binary feedback fails to assign credit to individual steps, leading to gradient signal collapse in failure dominant batches. To address these challenges, we introduce Step-Aware Contrastive Alignment (\textbf{SACA}), a framework designed to extract dense supervision from imperfect trajectories. At its core, the Perception-Grounded Step-Aware auditor evaluates progress step-by-step, disentangling failed trajectories into valid prefixes and exact divergence points. Leveraging these signals, Scenario-Conditioned Group Construction mechanism dynamically routes batches to specialized resampling and optimization strategies.  Extensive experiments on VLN-CE benchmarks demonstrate that SACA achieves state-of-the-art performance.

\keywords{Vision-Language Navigation \and Multi-modal Large Language Models \and Reinforcement Fine-Tuning \and Embodied Intelligence}
\end{abstract}

%% file: sections/Introduction.tex
\section{Introduction}
\label{sec:intro}

\begin{wrapfigure}{r}{0.5\textwidth}
    \centering
    \includegraphics[width=0.5\textwidth]{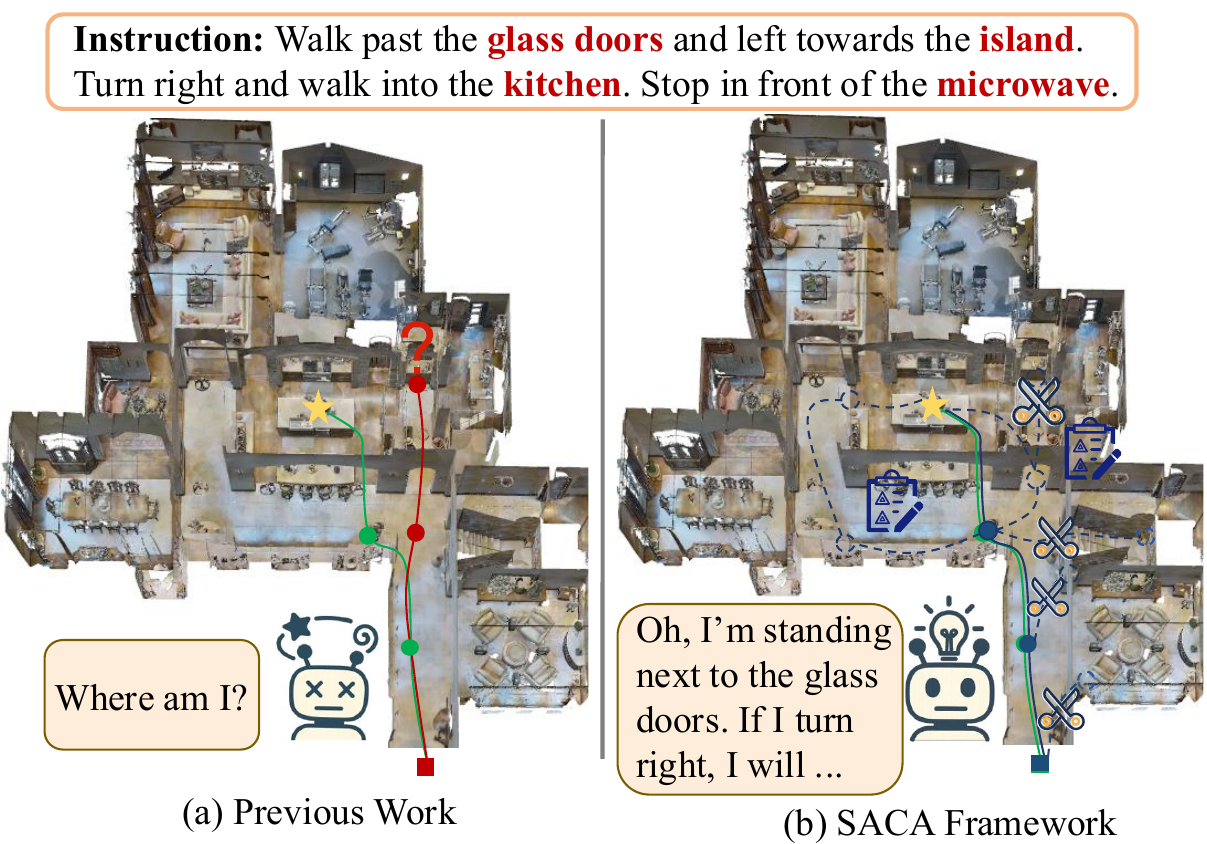}
    \vspace{-1em}
    \caption{\textbf{(a)} Previous work discard entire trajectory upon failure due to compounding errors and sparse rewards. \textbf{(b)} SACA uses the PGSA auditor to pinpoint the exact divergence point (\inlineicon{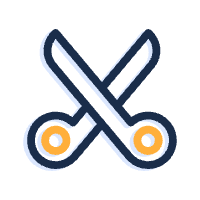}{1.3em}{-0.5em}). Then uses Repair Resampling (\inlineicon{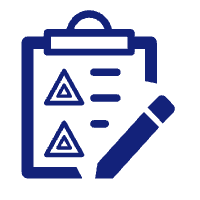}{1.3em}{-0.5em}) to recover from near-miss trajectories.}
    \label{fig:first}
    \vspace{-2em}
\end{wrapfigure}

Vision-Language Navigation in Continuous Environments (VLN-CE) requires agents to interpret natural language instructions, process visual streams and execute low-level actions~\cite{anderson2018r2r, ku2020rxr}. While Video-based Large Language Models (Video-LLMs) have improved spatial-temporal reasoning and serve as promising foundation models, applying them to continuous navigation is challenging due to trade-offs among generalization, error recovery and training stability~\cite{wei2026streamvln, qi2025vln}. The standard adaptation approach for VLN relies on Supervised Fine-Tuning (SFT) using expert data~\cite{wei2026streamvln, cheng2024navila, zheng2024navillm}. Although SFT quickly aligns models' behavior with instructions, policies trained purely by imitation suffer from compounding errors. As shown in Fig.~\ref{fig:first}, minor deviations push agents into out-of-distribution (OOD) states where policies trained by SFT often fail~\cite{qi2025vln}. 

Reinforcement Fine-Tuning (RFT) addresses this by allowing autonomous exploration. Group Relative Policy Optimization (GRPO)~\cite{grpo, Deepseek-R1} is a highly efficient RFT algorithm that eliminates the need for expensive value functions and critical models. However, standard GRPO struggles in VLN-CE due to sparse navigation rewards. The environment typically provides binary feedback only when the agent executes a \texttt{STOP} action. This coarse signal fails at step-level credit assignment, treating immediate failures and near-misses the same. During early exploration, this phenomenon often results in batches where all trajectories fail. In these all-failure groups, GRPO's relative advantage vanishes, leading to gradient collapse and wasted computation. While Process Reward Models (PRMs)~\cite{yin2025dynamic,cui2025process,wu2023fine} can provide dense supervision, training domain-specific PRMs is costly and prone to reward hacking.

We observe that not all failures are equally uninformative. In fact, our experimental data indicates that approximately \textbf{73\%} of failed episodes successfully execute the initial sub-instructions, yielding valuable valid prefixes that standard RL navely discards (detail in Appendix~\ref{app:statistics}). To extract this dense, step-level supervision from imperfect trajectories without relying on expensive, domain-specific PRMs, we propose Step-Aware Contrastive Alignment (SACA). At its core, SACA employs Perception-Grounded Step-Aware (PGSA) auditor that evaluates progress step-by-step against landmarks extracted from instructions. Integrating precise spatial grounding with semantic alignment, the PGSA auditor provides a robust continuous Soft Score for trajectory ranking alongside a discrete Structural Mask. This mask explicitly disentangles the trajectory into a Valid Prefix and a Divergence Point, enabling SACA to pinpoint the earliest moment the agent drifted off-course.

Leveraging these granular signals, SACA employs a Scenario-Conditioned Group Construction mechanism to dynamically route batches to specialized optimization strategies. If a sampled group has at least one success (mixed-outcome), outcome rewards drive training: near-miss failures are utilized through Repair Resampling, replanning from the Divergence Point to synthesize corrective demonstrations. Conversely, if all trajectories fail (null-outcome), SACA triggers All-Failure Rescue. It forms a Reflection Sub group using the best failure (the Pseudo-Anchor) alongside mined hard negatives, restoring relative supervision via conservative process-aware advantages. Furthermore, fine-grained step-level constraints are applied exclusively to the Pseudo-Anchor: Consistency Alignment reinforces the Valid Prefix through behavior cloning, while Contrastive Correction explicitly penalizes the Divergence Point. These mechanisms effectively convert failed rollouts into trajectory- and step-level supervision.

In summary, the main contributions are three-fold: \textbf{(i)} We propose Step-Aware Contrastive Alignment (SACA), a framework that resolves learning-signal collapse in sparse-reward by extracting dense supervision from imperfect trajectories. This is achieved through the Perception-Grounded Step-Aware (PGSA) auditor that leverages zero-shot foundation models for precise spatial and semantic tracking, bypassing the need for domain-specific PRMs ($\bm{\S}$\ref{subsec:Auditor}). \textbf{(ii)} We introduce a Scenario-Conditioned Group Construction mechanism that dynamically switches between Repair Resampling for mixed-outcome groups and All-Failure Rescue for null-outcome groups ($\bm{\S}$\ref{subsec:Group Construction}). We further propose the robust SACA optimization objective that combines trajectory-level advantages with step-level Consistency Alignment and Contrastive Correction ($\bm{\S}$\ref{subsec:Optimization Objective}). \textbf{(iii)} Extensive experiments on VLN-CE benchmarks demonstrate that SACA achieves state-of-the-art performance, providing an efficient exploration paradigm for continuous embodied tasks ($\bm{\S}$\ref{subsec:comparisons}).

%% file: sections/Related_work.tex
\section{Related Work}
\label{sec:related_work}

\noindent \textbf{Vision-Language Navigation in Continuous Environments (VLN-CE).} Vision-Language Navigation (VLN) requires embodied agents to execute natural language instructions~\cite{anderson2018r2r,qi2020reverie}. While early research focused on discrete topological graphs~\cite{liu2023bird,liu2024volumetric,liu2024vision,fan2024navigation,fan2024scene,gao20253d,gao2026uncertainty,chen2022duet,Wang_2021_scm,zhou2024navgpt}, recent paradigms have shifted to continuous environments (VLN-CE)~\cite{raychaudhuri2021law,georgakis2022cross,chen2022weakly,an2023etpnav,Wang_2023_dreamwalker,krantz_vlnce_2020}, demanding fine-grained low-level control. To manage this complexity, many methods rely on simulator-pretrained waypoint predictors~\cite{an2023etpnav,wang2023scaling,Wang_2024_lookahead}. However, these predictors often overfit training layouts, bottlenecking generalization to unseen scenes and highlighting the need for more adaptable, long-horizon navigation architectures.

\noindent \textbf{Navigation with Multi-modal Large Language Models (MLLMs).} The rapid evolution of MLLMs has unlocked new avenues for VLN. Some approaches~\cite{zhou2024navgpt,long2023discuss,chen20232a2nav,long2024instructnav} integrate MLLMs as zero-shot high-level planners to achieve instruction annotation with high linguistic diversity, though they typically underperform task-specific models. Conversely, another line of research~\cite{zhang2024navid,zhang2024uninavid,cheng2024navila,zhang2025mapnav} focuses on end-to-end Supervised Fine-Tuning (SFT) of Video-LLMs~\cite{zhang2024llavavideo,lin2024vila,li2024llamavid, wei2026streamvln} to directly parse spatio-temporal dynamics and output low-level control commands. 

\noindent \textbf{Reinforcement Fine-Tuning (RFT) in Embodied Intelligence.} RFT can enhance the ability of LLMs and MLLMs in reasoning and generalization~\cite{Deepseek-v3, Deepseek-R1, grpo, hurst2024gpt, jaech2024openai, singh2025openai, zhai2024fine}. To balance alignment with computational efficiency and learning complexity, RFT has rapidly evolved from DPO~\cite{rafailov2023direct} and PPO~\cite{schulman2017proximal} to highly scalable, critic-free frameworks \textit{e.g.} GRPO~\cite{grpo, Deepseek-R1} and GSPO~\cite{zheng2025group}. They span several paradigms, including offline, online, and test-time RL. While RFT was a common technique in Physics-based Simulator~\cite{liang2018gpu, koenig2004design},  VLA~\cite{guo2025improving, zhang2025embodied, guo2025vla, ye2025vla} and discrete VLN~\cite{wang2018look, chen2022reinforced}, its direct application to VLN-CE is imcomplete due to the expanded action space and longer trajectory length.  Recently, VLN-R1~\cite{qi2025vln} pioneered GRPO in VLN-CE. Yet, it remains bottlenecked by sparse outcome supervision. Proposed SACA framework introduces the PGSA auditor to extract dense, step-level signals from imperfect trajectories. Leveraging these signals, Scenario-Conditioned Group Construction mechanism dynamically reconstructs pseudo-positive samples from failures. By structurally isolating valid prefixes and repairing divergence points, SACA provides continuous supervision, effectively overcoming the sample inefficiency of sparse rewards.

%% file: sections/Method.tex
\section{Method}
\label{sec:method}
In this section, we present the SACA framework. We first introduce the PGSA auditor, which extracts continuous ranking signals and discrete structural boundaries from uninformative rollouts without requiring a pre-trained reward model ($\bm{\S}$\ref{subsec:Auditor}). Then, we detail the Scenario-Conditioned Group Construction mechanism, dynamically switching between Repair Resampling for mixed-outcome groups and All-Failure Rescue for null-outcome groups ($\bm{\S}$\ref{subsec:Group Construction}). Finally, we formulate the robust SACA optimization objective, integrating trajectory-level conservative advantages with step-level corrective constraints ($\bm{\S}$\ref{subsec:Optimization Objective}).

\begin{figure}[t]
    \vspace{-1em}
    \centering
    \includegraphics[width=\linewidth]{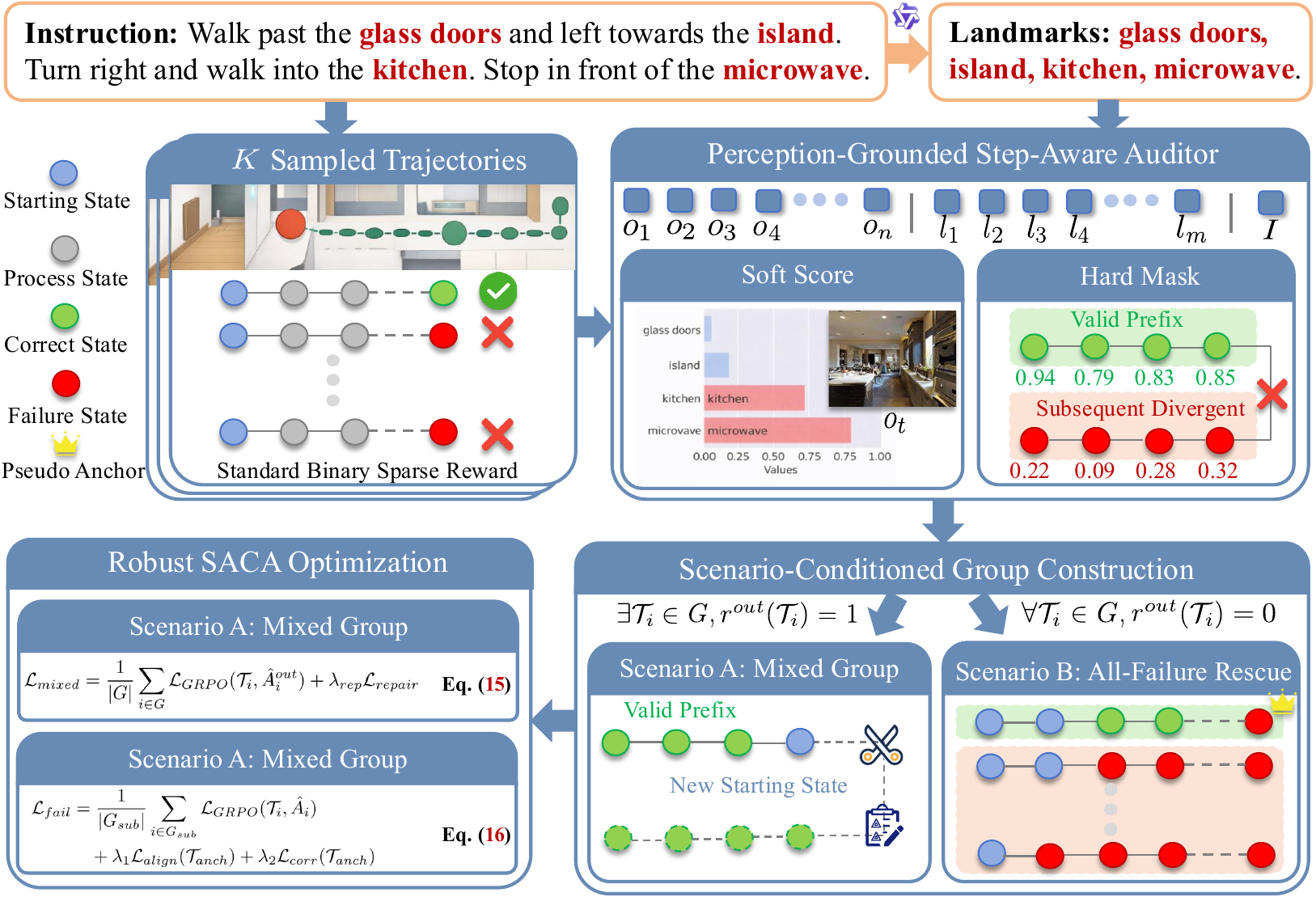}
    \vspace{-1em}
    \caption{\textbf{Overview of proposed SACA framework.} The PGSA auditor evaluates $K$ trajectories against instruction landmarks, yielding a Soft Score for ranking and a Hard Mask to isolate the Divergence Point. Based on batch outcomes, a Scenario-Conditioned mechanism dynamically routes to either Repair Resampling (for mixed groups) or All-Failure Rescue (for null-outcome groups), followed by robust optimization.}
    \label{fig:method}
    \vspace{-1.5em}
\end{figure}

\vspace{-1em}
\subsection{Perception-Grounded Step-Aware (PGSA) Auditor}
\label{subsec:Auditor}
\vspace{-0.5em}

To bridge the semantic gap between high-level linguistic instructions $I$ and low-level continuous observations $O=\{o_1, \dots,o_T\}$, the PGSA auditor employs a hierachical, zero-shot perception pipeline. Given an instruction, we first use a frozen tiny LLM (\textit{e.g.}, Qwen3-0.6B~\cite{yang2025qwen3}) to parse into a sequence of intermediate landmarks $L = \{l_1, \dots, l_m\}$. By synergizing foundation models (\textit{e.g.}, GroundingDINO~\cite{liu2024grounding}, SAM3~\cite{carion2025sam3segmentconcepts}, CLIP~\cite{radford2021learning}), the PGSA auditor synthesizes two complementary supervision signals: a continuous \texttt{soft score} for fine-grained trajectory ranking, and a discrete \texttt{hard mask} to structurally isolate the exact divergence point where the policy deviates from the optimal path.

\noindent \textbf{Hierarchical Soft Scoring.} To formulate a robust evaluation of progress toward a target landmark $l_i$, we construct a composite soft score $S_t$ that progressively fuses global contextual awareness with precise object-level grounding. Specifically, the PGSA auditor first computes a global semantic similarity via CLIP~\cite{radford2021learning}. GroundingDINO~\cite{liu2024grounding} acts as a spatial bottleneck, predicting a bounding box and an associated detection confidence $c_{det}$. To distill pure object-centric semantics and filter out background noise, instances with high detection confidence ($c_{det} \ge \tau_{det}$) are further refined by SAM3~\cite{carion2025sam3segmentconcepts}, yielding a precise object mask $M_{obj}$ with an IoU score $c_{iou}$. The integrated reward is formulated as:
\vspace{-0.5em}
\begin{equation}
\label{eq:soft_score}
    S_t(o_t, l_i) = w_1 \text{sim}(o_t, l_i) + \mathbb{I}(c_{det} \ge \tau_{det}) \big[ w_2 c_{det} + w_3 (c_{iou} \cdot \text{sim}(o_t \odot M_{obj}, l_i)) \big],
\end{equation}
\vspace{-1.5em}

\noindent where $w_1, w_2, w_3$ are balancing weights. This progressive formulation ensures that $S_t$ gracefully relies on global context when the target is distant, yet exhibits a sharp, highly informative peak upon precise spatial grounding, providing the agent with dense, low-variance positive reinforcement.

To derive a scale-invariant trajectory-level evaluation and mitigate the length bias inherent in cumulative step-wise rewards, we aggregate the step scores into a normalized process score $R_{proc}(\mathcal{T})$ via a threshold-gated formulation:
\vspace{-0.5em}
\begin{equation}
\label{eq:r_proc}
    R_{proc}(\mathcal{T}) = \frac{1}{T} \sum_{t=1}^{T} \text{ReLU}(S_t - \tau_s),
\end{equation} 
\vspace{-1em}

\noindent where $T$ represents the trajectory length. Here, the ReLU function acts as a noise-filtering gate: it truncates low-confidence step scores below the threshold ($\tau_s$) to zero, ensuring that only valid perception steps strongly aligned with the instruction can accumulate and contribute to the final trajectory score. Please refer to Appendix~\ref{app:more_details_SACA} for detailed visualizations of the step-level trajectory score. 

\noindent \textbf{Structural Hard Masking.} Simultaneously, we apply a stricter hard threshold $\tau_h$ to generate a step-wise binary mask $M_t= \mathbb{I}(S_t > \tau_h)\in \{0, 1\}$. During navigation, the active landmark $l_i$ advances to the next landmark in $L$ once this hard-match condition is satisfied for $c$ consecutive steps. This mask structurally disentangles the trajectory by identifying the exact step where the agent deviates from the instruction. We define the Divergence Point $t_{div}$ as:
\vspace{-0.5em}
\begin{equation}
\label{eq:t_div}
    t_{div} = 
    \begin{cases} 
        \min \{t \in \{1, \dots, T\} \mid M_t = 0\}, & \text{if } \exists t \text{ s.t. } M_t = 0, \\ 
        T + 1, & \text{otherwise}.
    \end{cases}
\end{equation}
\vspace{-1em}

\begin{figure}[t]
    \centering
    \includegraphics[width=\textwidth]{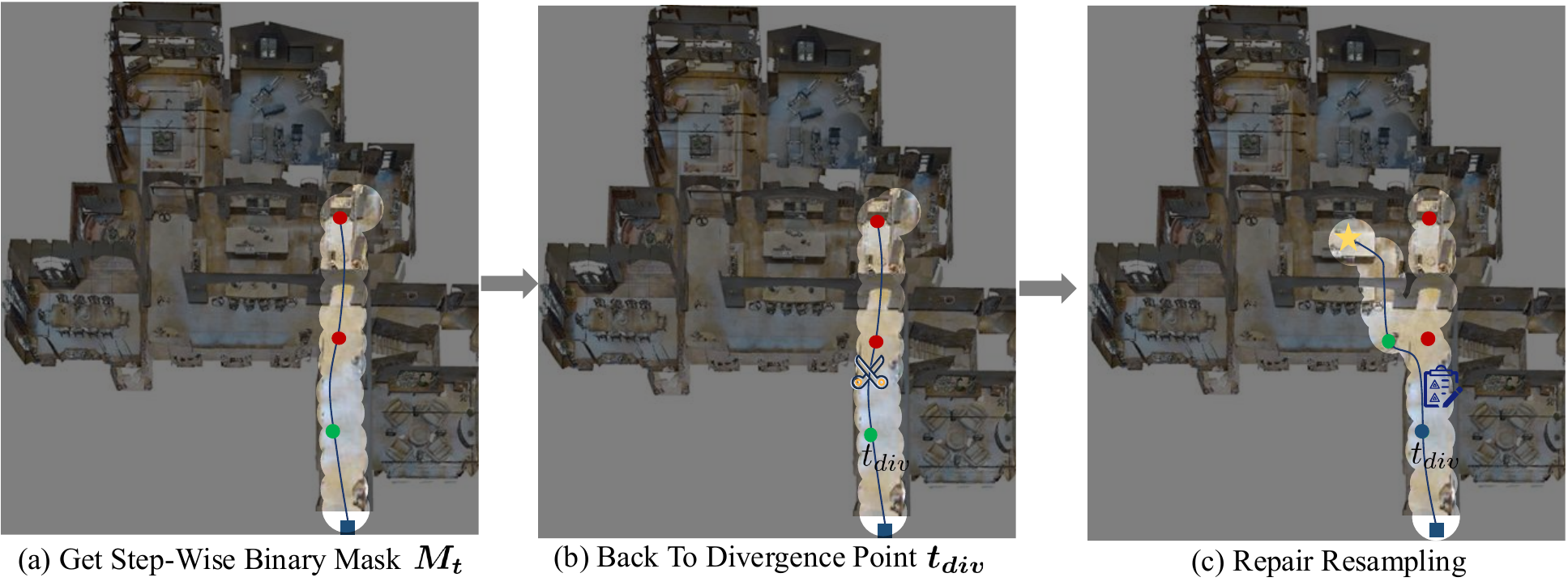}
    \caption{\textbf{Illustraion of the Repair Resampling process.} \textbf{(a)} Extracting the structural mask $M_t$ via the PGSA auditor. \textbf{(b)} Backtracking to the Divergence Point $t_{div}$ to prune the erroneous suffix. \textbf{(c)} Resampling a corrective path from $t_{div}$, thereby salvaging the valid prefix and providing robust step-level supervision.}
    \label{fig:t_div}
    \vspace{-2em}
\end{figure}

As shown in Fig.~\ref{fig:t_div}, the trajectory is partitioned into a Valid Prefix comprising all steps $t < t_{div}$, and a subsequent divergent phase starting at $t_{div}$. This decoupling enables smooth trajectory-level ranking via $R_{proc}$ while preserving strict discrete boundaries for step-level correction.

\vspace{-1em}
\subsection{Scenario-Conditioned Group Construction Mechanism}
\label{subsec:Group Construction}
\vspace{-0.5em}

Let $r^{out}(\mathcal{T}) \in \{0, 1\}$ denote the binary outcome reward. For each instruction, the policy model $\pi_{\theta}$ samples a group of trajectories: $G = \{\mathcal{T}_1, \dots, \mathcal{T}_K\}$. The core principle is using outcome supervision whenever success exists.

\noindent \textbf{Scenario A: Mixed Group ($\exists \tau_i \in G, r^{out}(\tau_i) = 1$).} When $G$ contains successful trajectories, outcome supervision drives the primary optimization. To further boost sample efficiency, we apply \textbf{Repair Resampling} to high-quality ``near-miss'' failures. We formally define this near-miss subset as trajectories:
\vspace{-0.5em}
\begin{equation}
    G_{miss} = \left\{ \mathcal{T}_i \in G \mid r^{out}(\mathcal{T}_i) = 0 \land \frac{t_{div}^{(i)}}{T_i} > \eta \right\},
\end{equation}
\vspace{-1em}

\noindent where the valid-prefix ratio exceeds a threshold $\eta$. For each $\mathcal{T}_i \in G_{miss}$, we truncate the rollout at the Divergence Point $t_{div}^{(i)}$ and iteratively resample suffixes $\mathcal{T}_{suff}^{(n)} \sim \pi_{\theta}(\cdot \mid \mathcal{T}_{i, 1:t_{div}^{(i)}})$ up to $N_{rep}$ times. Then halt resampling and construct a repaired trajectory via concatenation:
\vspace{-0.5em}
\begin{equation}
    \mathcal{T}_i^{rep} = \mathcal{T}_{i, 1:t_{div}^{(i)}} \oplus \mathcal{T}_{suff}^{(n)}.
\end{equation}
\vspace{-1em}

These successfully synthesized trajectories form an auxiliary set $\mathcal{T}^{rep}$ to augment the current training update, strictly without recursive resampling. If all $N_{rep}$ attempts fail, the original $\mathcal{T}_i$ is retained.

\noindent \textbf{Scenario B: All-Failure Rescue ($\forall \mathcal{T}_i \in G, r^{out}(\mathcal{T}_i) = 0$).} When all $K$ trajectories fail, standard outcome-based GRPO collapses. SACA triggers the \textbf{All-Failure Rescue} by constructing a Reflection Sub-group $G_{sub}$ centered around a Pseudo-Anchor, which isolates the most informative failure and its nearby alternatives for corrective comparison. The Pseudo-Anchor $\mathcal{T}_{anch}$ is used only as the most informative failure, rather than a pseudo-positive target. First, we select the trajectory with the highest process score:
\vspace{-0.5em}
\begin{equation}
\label{eq:anchor}
    \mathcal{T}_{anch} = \argmax_{\tau \in G} R_{proc}(\mathcal{T}).
\end{equation} 
\vspace{-1em}

Next, we mine Hard Negatives using a dual-criterion score:
\vspace{-0.5em}
\begin{equation}
\label{eq:score_neg}
   S_{neg}(\mathcal{T}) = \lambda \cdot \text{PrefixSim}(\mathcal{T}, \mathcal{T}_{anch}) - (1-\lambda) \cdot (R_{proc}(\mathcal{T}_{anch}) - R_{proc}(\mathcal{T})),
\end{equation}
\vspace{-1.5em}

\noindent where we instantiate $\text{PrefixSim}$ using action-level Longest Common Subsequence (LCS). $\lambda$ is a balancing weight. Both Hard-Negative mining and Repair Resampling are centered on the same structural prior: informative failures tend to share a correct prefix and diverge only in the later stage. We set $m < K$ and select the top-$m$ failed trajectories ranked by $S_{neg}$, excluding the Pseudo-Anchor itself, to construct the negative set $G_{neg}$, yielding $G_{sub} = \{\mathcal{T}_{anch}\} \cup G_{neg}$. $m$ is chosen such that $|G_{sub}| \ge 2$, ensuring stable subgroup normalization.

\vspace{-1em}
\subsection{Robust SACA Optimization Objective}
\label{subsec:Optimization Objective}
\vspace{-0.5em}

We use $\mathcal{L}_{GRPO}(\tau_i, \hat{A}_i)$ to denote the standard clipped policy-ratio loss evaluated on trajectory $\tau_i$ with advantage $\hat{A}_i$. Within the Reflection Sub-group $G_{sub}$, we compute the relative advantage to update the policy. To reduce the risk of wrongful penalization due to noisy visual-process estimates, we introduce a hierarchical conservative estimation mechanism:
\vspace{-0.5em}
\begin{equation}
\label{eq:a_raw}
    \hat{A}_i^{raw} = \frac{R_{proc}(\tau_i) - \mu\big(\{R_{proc}(\tau) \mid \tau \in G_{sub}\}\big)}{\sigma\big(\{R_{proc}(\tau) \mid \tau \in G_{sub}\}\big) + \epsilon_0}.
\end{equation} 
\vspace{-1em}

\noindent \textbf{Margin-Based Rescue.} First, we compute the margin $\Delta$:
\vspace{-0.5em}
\begin{equation}
\label{eq:margin}
    \Delta = R_{proc}(\mathcal{T}_{anch}) - \mu\big(\{R_{proc}(\mathcal{T}) \mid \mathcal{T} \in G_{neg}\}\big).
\end{equation}
\vspace{-1em}

If the Pseudo-Anchor lacks credibility ($\Delta < \delta$), we shrink the advantages by a temperature factor $\kappa < 1$:
\vspace{-1em}
\begin{equation}
\label{eq:a_prime}
    \hat{A}_i' = \begin{cases} \kappa \hat{A}_i^{raw}, & \text{if } \Delta < \delta, \\ \hat{A}_i^{raw}, & \text{otherwise}. \end{cases}
\end{equation}
\vspace{-1em}

\noindent \textbf{Negative-Only Scaling.} We apply an attenuation factor $s \in (0, 1]$ strictly to the negative advantages:
\vspace{-1em}
\begin{equation}
\label{eq:a_final}
    \hat{A}_i = \begin{cases} \hat{A}_i', & \hat{A}_i' \ge 0, \\ s \cdot \hat{A}_i', & \hat{A}_i' < 0. \end{cases}
\end{equation}
\vspace{-1em}

This preserves directional preference toward the most informative failure while reducing the risk of over-penalizing plausible alternatives under noisy visual estimates.

\noindent \textbf{Step-Level Constraints.} Beyond trajectory-level advantages, we apply fine-grained structural constraints exclusively to the Pseudo-Anchor $\mathcal{T}_{anch}$ to avoid amplifying noisy supervision from low-quality failures. We treat the actions along the Valid Prefix as pseudo-correct decisions and reinforce them via behavior cloning. By definition of $t_{div}$, all steps $t < t_{div}$ satisfy $M_t = 1$. Contrastive Correction is applied strictly to the Divergence Point $t_{div}$:

\begin{equation}
\label{eq:l_align}
    \mathcal{L}_{align}(\mathcal{T}_{anch}) = - \sum_{t < t_{div}} \log \pi_\theta(a_t \mid o_t, I),
\end{equation}
\vspace{-1em}
\begin{equation}
\label{eq:l_corr}
    \mathcal{L}_{corr}(\mathcal{T}_{anch}) = - \log \frac{\exp(\text{sim}(h_{s_{t_{div}}}, h_{a^+}) / \alpha)}{\exp(\text{sim}(h_{s_{t_{div}}}, h_{a^+}) / \alpha) + \exp(\text{sim}(h_{s_{t_{div}}}, h_{a^-}) / \alpha)},
\end{equation}
\vspace{-1em}

\noindent where $h_{s_{t_{div}}}$ denotes the policy's hidden state at the Divergence Point $t_{div}$. During training, $a^+$ is instantiated as a local teacher action derived from the simulator's shortest-path signal and is used exclusively for corrective supervision.

\noindent \textbf{Summary.} SACA dynamically switches objectives based on the group scenario:
\begin{itemize}[leftmargin=*, label=\textbullet]
    \item For a Mixed Group, we use the outcome-based advantage $\hat{A}^{out}_i$ and the mixed strategy:
    \vspace{-0.5em}
    \begin{equation}
    \label{eq:a_out}
        \hat{A}^{out}_i = \frac{r^{out}(\mathcal{T}_i) - \mu(\{r^{out}(\mathcal{T}) \mid \mathcal{T} \in G\})}{\sigma(\{r^{out}(\mathcal{T}) \mid \mathcal{T} \in G\}) + \epsilon_0},
    \end{equation}
    \begin{equation}
    \label{eq:l_mixed}
        \mathcal{L}_{mixed} = \frac{1}{|G|} \sum_{i \in G} \mathcal{L}_{GRPO}(\mathcal{T}_i, \hat{A}^{out}_i) - \lambda_{rep} \frac{1}{|\mathcal{T}_{rep}|} \sum_{\mathcal{T} \in \mathcal{T}_{rep}} \sum_{t \in \text{suffix}(\mathcal{T})} \log \pi_\theta(a_t \mid o_t, I),
    \end{equation}
    
     where $\lambda_{rep}, \epsilon_0$ are balancing weights. 

    \item For an All-Failure Group, we utilize the robust process advantage $\hat{A}_i$ alongside the anchor-specific constraints:
    
    \begin{equation}
    \label{eq:l_fail}
        \mathcal{L}_{fail} = \frac{1}{|G_{sub}|} \sum_{i \in G_{sub}} \mathcal{L}_{GRPO}(\mathcal{T}_i, \hat{A}_i) + \lambda_1 \mathcal{L}_{align}(\mathcal{T}_{anch}) + \lambda_2 \mathcal{L}_{corr}(\mathcal{T}_{anch}),
    \end{equation}
    where $\lambda_1, \lambda_2$ are balancing weights. 
   
\end{itemize}

In each training iteration, SACA first scores sampled trajectories with the PGSA  auditor, then routes the group to either Mixed-Group Repair or All-Failure Rescue based on outcome availability, and finally updates the policy with scenario-specific objectives. Every sampled rollout is assigned a usable training role \textit{e.g.} successful, repairable, or rescuable. Even failed trajectories contribute structured supervision under severe reward sparsity. Additional step-by-step explanation of the SACA framework is provided in Appendix~\ref{app:more_details_SACA}.

%% file: sections/Experiments.tex
\section{Experiment}
In this section, we comprehensively evaluate the SACA framework. We first outline the experimental setup, detailing the VLN-CE benchmarks, evaluation metrics, and implementation details ($\bm{\S}$\ref{subsec:setup}). Building upon this foundation, we compare SACA with existing state-of-the-art methods, demonstrating its superior generalization and robust error-recovery capabilities in both standard and complex long-horizon navigation tasks ($\bm{\S}$\ref{subsec:comparisons}). Finally, we present extensive ablation studies to validate the individual contributions of the PGSA perception modules, the Scenario-Conditioned Group Construction mechanism, and the robust SACA optimization objective ($\bm{\S}$\ref{subsec:ablation}).
\input{tables/tab_main}

\subsection{Experimental Setup}
\label{subsec:setup}

\noindent \textbf{Datasets and Evaluation.} We evaluate the SACA framework on two prominent Vision-Language Navigation in Continuous Environments (VLN-CE) benchmarks: R2R-CE~\cite{anderson2018r2r} and RxR-CE~\cite{ku2020rxr}. Both datasets are constructed over the Matterport3D (MP3D) ~\cite{Matterport3D} scenes and simulated within the Habitat platform~\cite{savva2019habitat,ramakrishnan2021habitat, puig2023habitat3}. Following the standard evaluation protocol established by previous VLN-CE works, we assess navigation performance using a comprehensive suite of metrics. These include Trajectory Length (TL), Navigation Error (NE), Oracle Success Rate (OS), Success Rate (SR), and Success weighted by Path Length (SPL). For RxR-CE, we additionally report normalized Dynamic Time Warping (nDTW) to measure the spatial fidelity between the predicted and ground-truth paths. Among these metrics, SR and SPL on the unseen splits remain the most critical indicators of an agent's generalization capability.

\noindent \textbf{Implementation Details.} We initialize the policy network using a pre-trained Video-LLM (\textit{e.g.}, LLaVA-Video-8B~\cite{zhang2024llavavideo}) and first perform SFT to align its visual representations with basic navigation instructions, deployed on 8 Nvidia A6000 GPUs. During SFT phase, we employ a learning rate of $1 \times 10^{-5}$ with a cosine schedule (10\% warmup), which requires approximately 36 hours. Subsequently, during the RFT phase, we reduce the learning rate to $1 \times 10^{-6}$ with a weight decay of 0.01 and set the KL penalty coefficient $\beta = 0.04$, completing one epoch in about 24 hours. For a comprehensive list of hyperparameters and detailed training configurations, please refer to the Appendix~\ref{app:hyperparams}.

\begin{figure}[t]
    \centering
    \includegraphics[width=\linewidth]{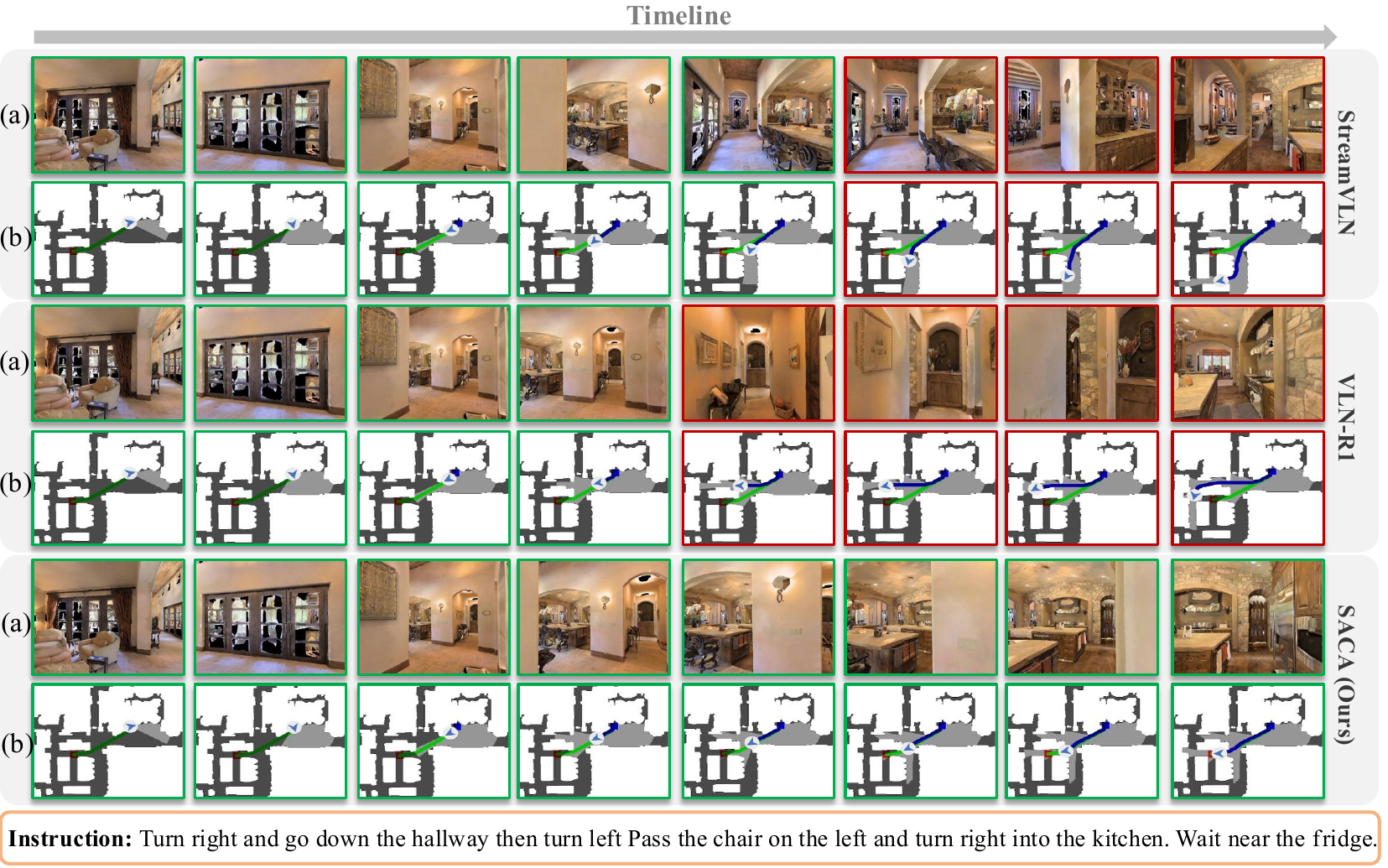}
    \caption{\textbf{Qualitative comparison of navigation trajectories.} \textbf{(a)} Egocentric observations. \textbf{(b)} Top-down maps. Green and red frames denote correct trajectory and failure trajectory, respectively.}
    \vspace{-2em}
\label{fig:qualitative_comp}
\end{figure}

\subsection{Comparisons with State-of-the-Arts}
\label{subsec:comparisons}

Table~\ref{tab:comp-vlnce} compares SACA with state-of-the-art methods on the val-unseen splits of VLN-CE R2R~\cite{anderson2018r2r} and RxR~\cite{ku2020rxr}. Baselines are categorized by observation encoders and extra data usage. SACA establishes a new state-of-the-art across nearly all metrics. We also provide visualizations of comparison experiments in Fig.~\ref{fig:qualitative_comp}. While StreamVLN~\cite{wei2026streamvln} and VLN-R1~\cite{qi2025vln} fail to recover from deviations, SACA successfully maintains step-level alignment to reach the destination.

\input{tables/tab_ablation_components}

\noindent \textbf{Superiority in Standard Ego-view Settings.} Compared to recent single-RGB methods (\textit{e.g.}, VLN-R1~\cite{qi2025vln}, StreamVLN~\cite{wei2026streamvln}) using no extra data, SACA achieves \textbf{60.3\% SR} and \textbf{55.1\% SPL} on R2R-CE. This yields substantial absolute improvements of \textbf{7.5\%} (SR) and \textbf{7.9\%} (SPL) over the previous best, StreamVLN, while reducing Navigation Error (NE) to 4.57. This validates the profound efficacy of mining step-level supervision from failed rollouts.

\noindent \textbf{Robustness in Long-Horizon Navigation.} On the challenging, long-horizon RxR-CE benchmark, SACA's advantages magnify. It attains \textbf{60.3\% SR} and \textbf{49.8\% SPL}, surpassing the previous SOTA by massive margins of \textbf{11.7\%} (SR) and \textbf{7.3\%} (SPL). By structurally breaking down credit assignment, the PGSA auditor effectively mitigates the gradient collapse typical of standard methods.

\noindent \textbf{Scaling with Additional Data.} When augmented with the ScaleVLN~\cite{wei2026streamvln} dataset ($\dagger$), SACA further improves to \textbf{62.7\% SR} (R2R-CE) and \textbf{62.1\% SR} (RxR-CE). Notably, SACA$\dagger$ outperforms StreamVLN$\dagger$ by \textbf{9.2\%} in RxR SR.

\noindent \textbf{Surpassing Privileged Modalities.} Relying strictly on single ego-centric RGB images, SACA significantly outperforms methods heavily engineered for multi-sensor fusion. It surpasses ETPNav (which requires Panoramic views, Odometry, and Depth modalities) by \textbf{3.3\%} and \textbf{5.6\%} in R2R and RxR SR, respectively. This suggests that proposed dense, step-aware RL signals enable MLLMs to implicitly build spatial awareness that eclipses explicit multi-sensor mapping.


\subsection{Ablation Study}
\label{subsec:ablation}

\begin{wrapfigure}{r}{0.5\textwidth}
    \vspace{-2em}
    \centering
    \includegraphics[width=0.5\textwidth]{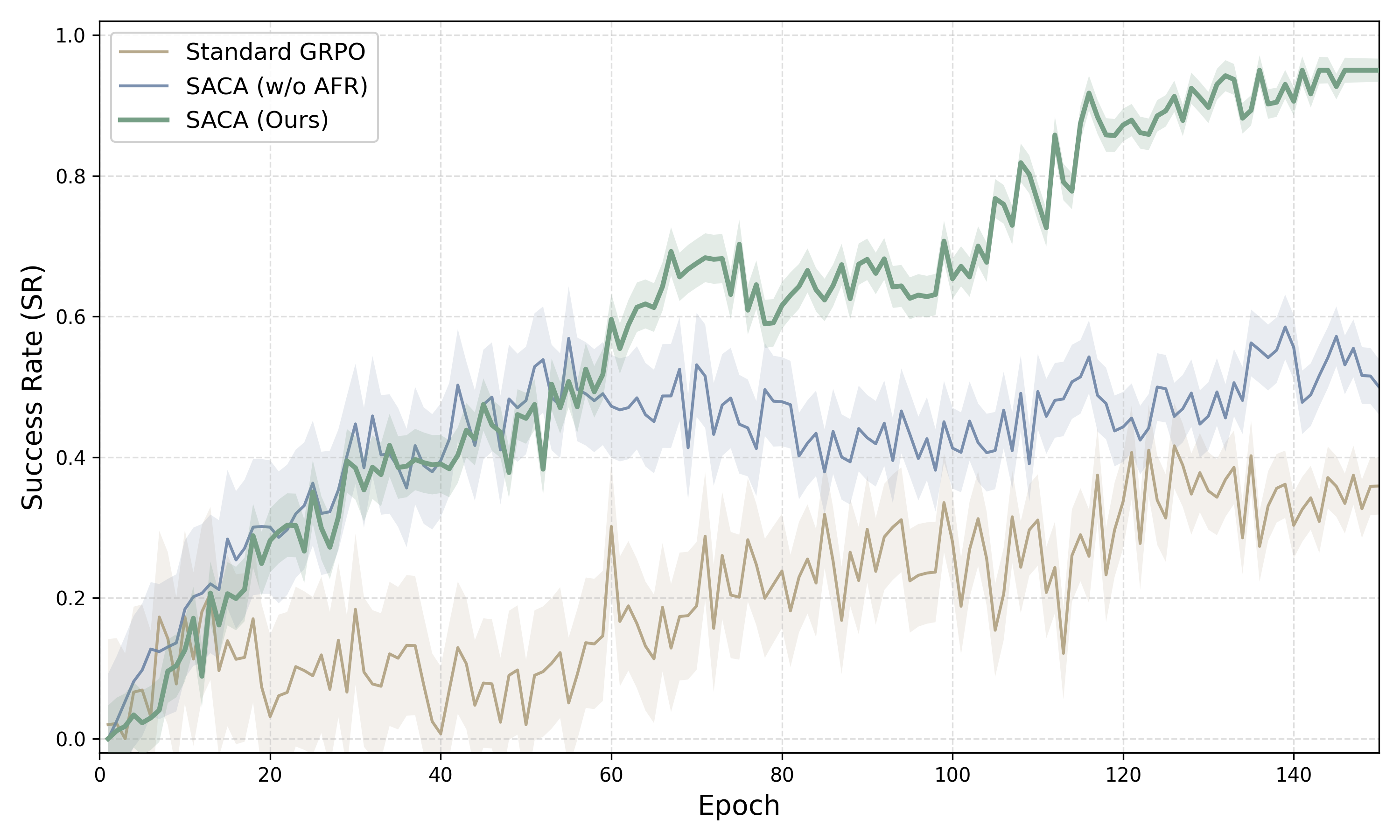}
    \caption{\textbf{Illustration of SR curves  during RFT training.}}
    \label{fig:reward}
    \vspace{-2em}
\end{wrapfigure}

We conduct extensive ablation studies on the VLN-CE R2R~\cite{anderson2018r2r} and RxR~\cite{ku2020rxr} Val-Unseen splits. The analysis is structured to validate the core macro-architecture (Table~\ref{tab:ablation_components} and Fig.~\ref{fig:reward}), the specific micro-objective designs (Table~\ref{tab:ablation_objectives}), and the hyperparameter sensitivity (Table~\ref{tab:ablation_hyperparams}).

\noindent \textbf{Training Dynamics and Sample Efficiency.} Fig.~\ref{fig:reward} illustrates the SR during RFT process. Standard GRPO (brown curve) suffers from severe learning-signal collapse and a premature plateau due to its reliance on sparse outcome rewards. Incorporating step-aware dense supervision without the All-Failure Rescue mechanism (SACA w/o AFR, blue curve) accelerates initial bootstrapping but eventually bottlenecks because it still discards null-outcome batches. In contrast, proposed full SACA framework (green curve) achieves vastly superior sample efficiency and asymptotic performance. By dynamically recovering dense supervision from completely failed batches via AFR, SACA maintains stable and effective gradient updates throughout training, successfully breaking the exploration bottleneck.

\begin{figure}[t]
    \vspace{-1em}
    \centering
    \includegraphics[width=\linewidth]{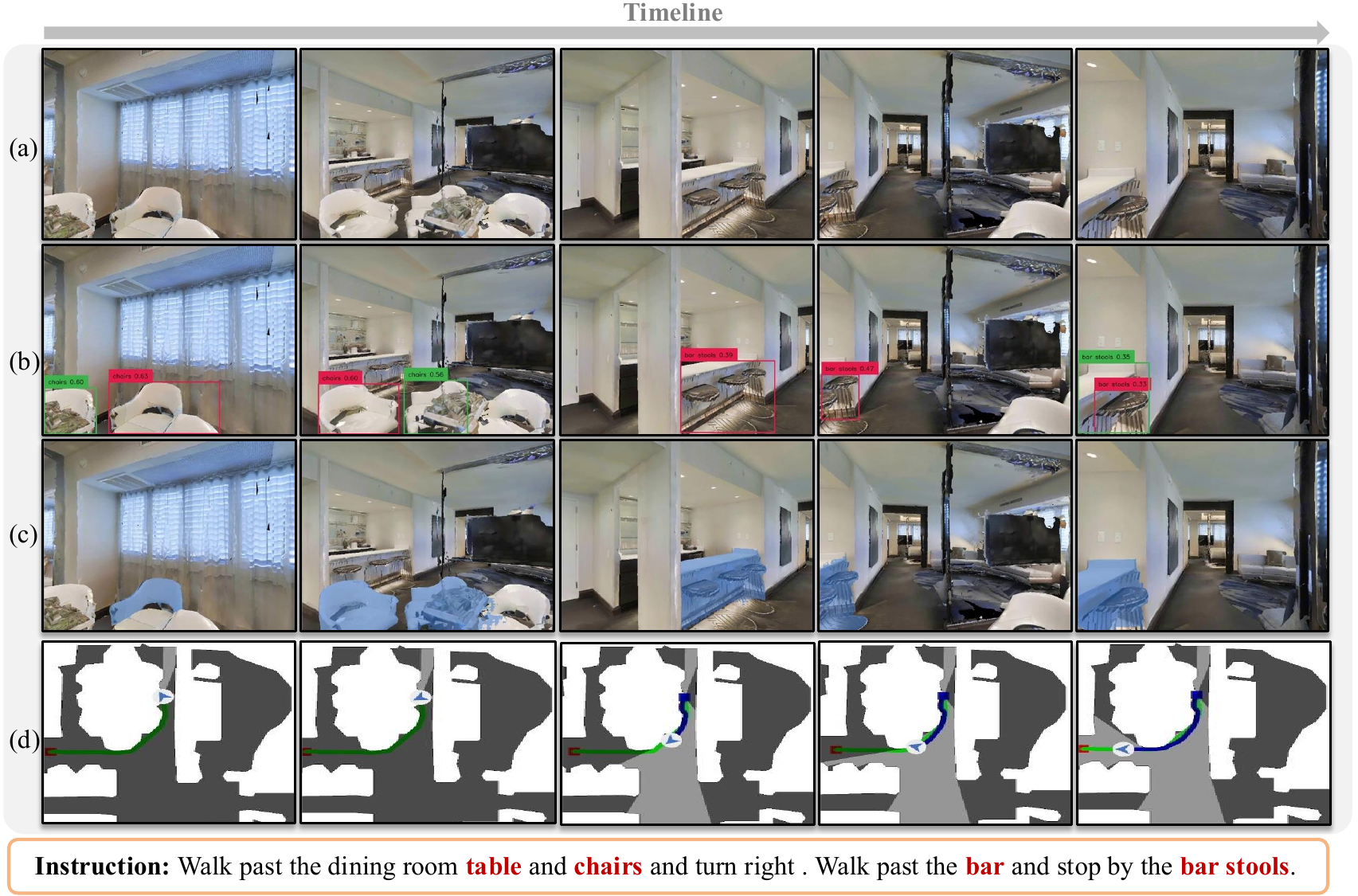}
    \caption{\textbf{Qualitative visualization of the Perception-Grounded Step-Aware (PGSA) Auditor's cascaded pipeline.}
    \textbf{(a)} Raw continuous egocentric observations $o_t$ during navigation. \textbf{(b)} GroundingDINO~\cite{liu2024grounding} detects intermediate landmarks. \textbf{(c)} SAM3~\cite{carion2025sam3segmentconcepts} extracts precise pixel-level masks. \textbf{(d)} The corresponding top-down trajectory mapping.}
\label{fig:pgsa_visualization}
\vspace{-2em}
\end{figure}

\noindent \textbf{Core Components Analysis.} Table~\ref{tab:ablation_components} demonstrates the progressive integration of SACA's core components starting from the SFT Baseline (Row 0). Relying exclusively on sparse outcome rewards, standard GRPO (Row 1) yields only marginal improvements (\textit{e.g.}, +1.3\% SR on RxR) and suffers from severe learning-signal collapse, as illustrated by the early plateau of the brown curve in Fig.~\ref{fig:reward}. Incorporating the continuous Soft Score (SS) from the PGSA auditor (Row 2) provides a dense ranking signal crucial for initial bootstrapping, steadily boosting the RxR SR to 51.8\%. The subsequent introduction of the All-Failure Rescue (AFR) mechanism (Row 3) brings substantial gains, driving RxR SR up by a massive 4.6\% to 56.4\%. As visually corroborated by Fig.~\ref{fig:reward}, AFR effectively recovers dense supervision from null-outcome batches (green curve), breaking the early performance bottleneck observed when AFR is disabled (blue curve). Finally, integrating Repair Resampling (Row 4) salvages near-miss failures by truncating trajectories at the exact Divergence Point and resampling the suffix. 

\input{tables/tab_ablation_objectives}
\noindent \textbf{Objective Constraints \& Robustness Mechanisms.} Table~\ref{tab:ablation_objectives} details the impact of the specific objectives and robust scaling factors. Removing Consistency Align ($\mathcal{L}_{align}$) significantly reduces navigation efficiency (\textit{e.g.}, R2R SPL drops from 55.1\% to 51.6\%), indicating that without behavior cloning on the valid prefix, the agent takes longer, sub-optimal paths. Conversely, omitting Contrastive Correct ($\mathcal{L}_{correct}$) heavily degrades long-horizon success (\textit{e.g.}, RxR SR drops to 56.4\%), emphasizing the necessity of explicitly penalizing the Divergence Point to prevent repeated mistakes at complex intersections. Furthermore, disabling robustness mechanisms like Margin-Based Rescue ($\kappa=1$) or Negative-Only Scaling ($s=1$) degrades overall performance. This demonstrates their crucial roles in filtering noisy gradients from low-confidence pseudo-anchors and acting as regularizers against the over-penalization of plausible alternative paths

\noindent \textbf{Qualitative Visualization of PGSA Auditor.} 
To intuitively demonstrate the efficacy of proposed cascaded perception modules, we visualize a navigation episode in Fig.~\ref{fig:pgsa_visualization}. Given an instruction containing intermediate landmarks (\textit{e.g.}, ``chairs'', ``bar stools''), raw visual observations (Row a) lack explicit object-level focus. GroundingDINO~\cite{liu2024grounding} (Row b) successfully detects these textual landmarks, yielding bounding boxes. However, standard bounding boxes inevitably encompass noisy background pixels. SAM3~\cite{carion2025sam3segmentconcepts} (Row c) refines these detections into precise pixel-level masks, strictly isolating the target objects. This pure spatial-semantic grounding ensures robust local CLIP~\cite{radford2021learning} alignment, allowing the auditor to accurately track progress on the top-down trajectory (Row d) and pinpoint the exact divergence point without any trained reward models.

\input{tables/tab_ablation_hyperparams}
\noindent \textbf{Hyperparameter Sensitivity.} Table~\ref{tab:ablation_hyperparams} evaluates SACA's sensitivity to group size ($K$) and maximum repair attempts ($N_{rep}$). A smaller group ($K=4$) suffers from high variance in advantage estimation (\textit{e.g.}, RxR SR drops to 55.2\%), while a larger group ($K=16$) introduces computational overhead and noisy long-tail trajectories that degrade performance (R2R SR drops to 60.1\%), making $K=8$ the optimal balance. For repair attempts, $N_{rep}=1$ provides insufficient exploration depth. Conversely, excessive attempts ($N_{rep}=5$) cause the policy to overfit to highly improbable, ``stitched'' trajectories, leading to a slight performance decay. Thus, $N_{rep}=3$ balances exploration depth and policy stability.

\input{tables/tab_ablation_pgsa}
\noindent \textbf{Perception Modules within PGSA auditor.} Table~\ref{tab:ablation_pgsa} evaluates the cascaded perception modules of the PGSA Auditor. Relying solely on global CLIP similarity (Row 1) yields suboptimal results (\textit{e.g.}, 57.6\% SR on R2R). Adding GroundingDINO~\cite{liu2024grounding} bounding boxes (Row 2) provides explicit object-level grounding, which increases R2R SR to 59.2\%. Finally, incorporating SAM3~\cite{carion2025sam3segmentconcepts} masks (Row 3) enables precise pixel-level segmentation. This crucial step filters out background noise to ensure pure local semantic alignment, ultimately pushing performance to the state-of-the-art 60.3\% SR on both datasets.

%% file: tables/tab_main.tex
\begin{table}[t]
\centering
\footnotesize
\setlength{\tabcolsep}{3pt}  
\label{tab:main}
\resizebox{\linewidth}{!}{
\begin{tabular}{l|cccc|cccc|cccc}
\toprule
\multirow{2}{*}{Method} & \multicolumn{4}{c|}{Observation Encoder} & \multicolumn{4}{c|}{R2R-CE Val-Unseen} & \multicolumn{4}{c}{RxR-CE Val-Unseen} \\ 
\cmidrule(lr){2-5} \cmidrule(lr){6-9} \cmidrule(lr){10-13}
      & Pano. & Odo. & Depth & S.RGB & NE$\downarrow$ & OS$\uparrow$ & SR$\uparrow$ & SPL$\uparrow$ & NE$\downarrow$ & SR$\uparrow$ & SPL$\uparrow$ & nDTW$\uparrow$ \\ 
\midrule
HPN+DN$^*$~\cite{krantz2021waypoint} & $\checkmark$ & $\checkmark$ & $\checkmark$ &  & 6.31  & 40.0  & 36.0  & 34.0  & - & - & - & - \\
CMA$^*$~\cite{hong2022bridging}      & $\checkmark$ & $\checkmark$ & $\checkmark$ &  & 6.20  & 52.0  & 41.0  & 36.0  & 8.76 & 26.5 & 22.1 & 47.0 \\
\vlnbert$^*$~\cite{hong2022bridging} & $\checkmark$ & $\checkmark$ & $\checkmark$ &  & 5.74  & 53.0  & 44.0  & 39.0  & 8.98 & 27.0 & 22.6 & 46.7 \\
Sim2Sim$^*$~\cite{krantz2022sim}    & $\checkmark$ & $\checkmark$ & $\checkmark$ &  & 6.07  & 52.0  & 43.0  & 36.0  & - & - & - & - \\
GridMM$^*$~\cite{wang2023gridmm}    & $\checkmark$ & $\checkmark$ & $\checkmark$ &  & 5.11  & 61.0  & 49.0  & 41.0  & - & - & - & - \\
ETPNav$^*$~\cite{an2023etpnav}      & $\checkmark$ & $\checkmark$ & $\checkmark$ &  & 4.71  & 65.0  & 57.0  & 49.0  & 5.64 & 54.7 & 44.8 & 61.9 \\ 
ScaleVLN$^{*}$~\cite{wang2023scaling} & $\checkmark$ & $\checkmark$ & $\checkmark$ &  & 4.80  & --    & 55.0  & 51.0  & -& - & - & -\\
\midrule
InstructNav~\cite{long2024instructnav} & - & - & - & - & 6.89 & --   & 31.0 & 24.0 & - & - & - & - \\
AG-CMTP~\cite{chen2021topological}   & $\checkmark$ & $\checkmark$ & $\checkmark$ &  & 7.90  & 39.2  & 23.1  & 19.1  & - & - & - & - \\
R2R-CMTP~\cite{chen2021topological}  & $\checkmark$ & $\checkmark$ & $\checkmark$ &  & 7.90  & 38.0  & 26.4  & 22.7  & - & - & - & - \\
LAW~\cite{raychaudhuri2021law}       &  & $\checkmark$ & $\checkmark$ & $\checkmark$ & 6.83  & 44.0  & 35.0  & 31.0  & 10.90 & 8.0 & 8.0 & 38.0 \\
CM2~\cite{georgakis2022cross}        &  & $\checkmark$ & $\checkmark$ & $\checkmark$ & 7.02  & 41.5  & 34.3  & 27.6  & - & - & - & - \\
WS-MGMap~\cite{chen2022weakly}       &  & $\checkmark$ & $\checkmark$ & $\checkmark$ & 6.28  & 47.6  & 38.9  & 34.3  & - & - & - & - \\
ETPNav + FF~\cite{Wang_2024_lookahead}  &  & $\checkmark$ & $\checkmark$ & $\checkmark$ & 5.95  & 55.8  & 44.9  & 30.4  & 8.79 & 25.5 & 18.1 & - \\
Seq2Seq~\cite{ku2020rxr}    &  &  & $\checkmark$ & $\checkmark$ & 7.77  & 37.0  & 25.0  & 22.0  & 12.10 & 13.9 & 11.9 & 30.8 \\
CMA~\cite{ku2020rxr}        &  &  & $\checkmark$ & $\checkmark$ & 7.37  & 40.0  & 32.0  & 30.0  & - & - & - & - \\
\midrule

VLN-R1~\cite{qi2025vln} &  &  &  & $\checkmark$ & 7.00 & 41.2 & 30.2 & 21.8 & 10.40 & 22.3 & 17.5 & - \\

NaVid~\cite{zhang2024navid}   &  &  &  & $\checkmark$ & 5.47  & 49.1  & 37.4  & 35.9  & - & - & - & - \\

MapNav~\cite{zhang2025mapnav}    &  &  &  & $\checkmark$ & 4.93  & 53.0  & 39.7  & 37.2  & - & - & - & - \\

NaVILA~\cite{cheng2024navila}   &  &  &  & $\checkmark$ & 5.37  & 57.6  & 49.7  & 45.5  & - & - & - & - \\

StreamVLN~\cite{wei2026streamvln}        &  &  &  & $\checkmark$ & 5.43  & 62.5  & 52.8  & 47.2  & 6.72 & 48.6 & 42.5 & 60.2 \\

\rowcolor{LightGray}
SACA (Ours) &  &  &  & $\checkmark$ & \textbf{4.57} & \textbf{64.9} & \textbf{60.3} & \textbf{55.1} & \textbf{4.90} & \textbf{60.3} & \textbf{49.8} & \textbf{62.1} \\

\midrule

NaVILA${\dagger}$~\cite{cheng2024navila}   &  &  &  & $\checkmark$ & 5.22  & 62.5  & 54.0  & 49.0  & 6.77 & 49.3 & 44.0 & 58.8 \\

UniNaVid${\dagger}$~\cite{zhang2024uninavid}   &  &  &  & $\checkmark$ & 5.58  & 53.3  & 47.0  & 42.7  & 6.24 & 48.7 & 40.9 & - \\

StreamVLN${\dagger}$~\cite{wei2026streamvln}   &  &  &  & $\checkmark$ & 4.98  & 64.2  & 56.9  & 51.9  & 6.22 & 52.9 & 46.0 & 61.9 \\

\rowcolor{LightGray}
SACA (Ours)${\dagger}$ &  &  &  & $\checkmark$ & \textbf{4.19} & \textbf{69.3} & \textbf{64.7} & \textbf{56.9} & \textbf{4.75} & \textbf{62.1} & \textbf{51.7} & \textbf{66.0} \\
\bottomrule
\end{tabular}}
\vspace{1mm}
\caption{\textbf{Comparison with state-of-the-art methods.} $*$ indicates methods using the waypoint predictor from~\cite{hong2022bridging}. $\dagger$ denotes methods using additional training data beyond the R2R-CE~\cite{anderson2018r2r} and RxR-CE~\cite{ku2020rxr} benchmarks.}
\label{tab:comp-vlnce}
\vspace{-3.5em}
\end{table}

%% file: tables/tab_ablation_components.tex
\begin{table}[t]
\centering
\footnotesize
\setlength{\tabcolsep}{4pt}
\caption{\textbf{Ablation study on the core components of SACA.} \textbf{SS}: Soft Score for ranking; \textbf{AFR}: All-Failure Rescue (Scenario B); \textbf{RR}: Repair Resampling (Scenario A).}
\label{tab:ablation_components}
\resizebox{\linewidth}{!}{
\begin{tabular}{c|ccc|cccc|cccc}
\toprule
\multirow{2}{*}{Row} & \multicolumn{3}{c|}{Components} & \multicolumn{4}{c|}{R2R-CE Val-Unseen} & \multicolumn{4}{c}{RxR-CE Val-Unseen} \\
\cmidrule(lr){2-4} \cmidrule(lr){5-8} \cmidrule(lr){9-12}
 & \textbf{SS} & \textbf{AFR} & \textbf{RR} & \textbf{NE}$\downarrow$ & \textbf{OS}$\uparrow$ & \textbf{SR}$\uparrow$ & \textbf{SPL}$\uparrow$ & \textbf{NE}$\downarrow$ & \textbf{SR}$\uparrow$ & \textbf{SPL}$\uparrow$ & \textbf{nDTW}$\uparrow$ \\
\midrule
0 & \multicolumn{3}{c|}{SFT Baseline~\cite{wei2026streamvln}} & 5.43 & 62.5 & 52.8 & 47.2 & 6.72 & 48.6 & 42.5 & 60.2 \\
\midrule
1 & & & & 5.39 & 55.3 & 54.1 & 48.8 & 6.54 & 49.9 & 43.0 & 60.3 \\ 
2 & $\checkmark$ & & & 5.21 & 63.2 & 55.4 & 49.6 & 6.15 & 51.8 & 44.3 & 60.7 \\ 
3 & $\checkmark$ & $\checkmark$ & & 4.82 & 64.1 & 58.2 & 52.4 & 5.48 & 56.4 & 47.6 & 61.3 \\ 
\rowcolor{LightGray}
4 & $\checkmark$ & $\checkmark$ & $\checkmark$ & \textbf{4.57} & \textbf{64.9} & \textbf{60.3} & \textbf{55.1} & \textbf{4.90} & \textbf{60.3} & \textbf{49.8} & \textbf{62.1} \\ 
\bottomrule
\end{tabular}}
\end{table}

%% file: tables/tab_ablation_objectives.tex
\begin{table}[t]
\centering
\footnotesize
\setlength{\tabcolsep}{4pt}
\caption{\textbf{Ablation study on specific objective designs and robustness mechanisms within SACA.} All variants are compared against the full SACA model.}
\label{tab:ablation_objectives}
\resizebox{\linewidth}{!}{
\begin{tabular}{l|cccc|cccc}
\toprule
\multirow{2}{*}{Model Variant} & \multicolumn{4}{c|}{R2R-CE Val-Unseen} & \multicolumn{4}{c}{RxR-CE Val-Unseen} \\
\cmidrule(lr){2-5} \cmidrule(lr){6-9}
 & \textbf{NE}$\downarrow$ & \textbf{OS}$\uparrow$ & \textbf{SR}$\uparrow$ & \textbf{SPL}$\uparrow$ & \textbf{NE}$\downarrow$ & \textbf{SR}$\uparrow$ & \textbf{SPL}$\uparrow$ & \textbf{nDTW}$\uparrow$ \\
\midrule
\rowcolor{LightGray}
\textbf{SACA} & \textbf{4.57} & \textbf{64.9} & \textbf{60.3} & \textbf{55.1} & \textbf{4.90} & \textbf{60.3} & \textbf{49.8} & \textbf{62.1} \\
\midrule
\multicolumn{9}{l}{\textit{Ablation on Objective Constraints}} \\
\quad w/o Consistency Align ($\mathcal{L}_{align}$) & 4.89 & 63.8 & 58.0 & 51.6 & 5.34 & 57.1 & 46.2 & 60.5 \\
\quad w/o Contrastive Correct ($\mathcal{L}_{correct}$) & 4.98 & 63.5 & 57.3 & 52.8 & 5.51 & 56.4 & 46.5 & 59.4 \\
\midrule
\multicolumn{9}{l}{\textit{Ablation on Robustness Mechanisms}} \\
\quad w/o Margin-Based Rescue ($\kappa=1$) & 4.85 & 64.1 & 58.5 & 53.0 & 5.25 & 57.8 & 47.9 & 61.2 \\
\quad w/o Negative-Only Scaling ($s=1$) & 4.76 & 64.5 & 59.1 & 54.0 & 5.12 & 58.6 & 48.5 & 61.6 \\
\bottomrule
\end{tabular}}
\end{table}

%% file: tables/tab_ablation_hyperparams.tex
\begin{table}[t]
\centering
\footnotesize
\setlength{\tabcolsep}{3pt}
\caption{\textbf{Ablation study on the group size ($K$) and maximum repair attempts ($N_{rep}$).}}
\label{tab:ablation_hyperparams}
\resizebox{\linewidth}{!}{
\begin{tabular}{cc|cccc|cccc}
\toprule
\multirow{2}{*}{\textbf{Group Size ($K$)}} & \multirow{2}{*}{\textbf{Max Repairs ($N_{rep}$)}} & \multicolumn{4}{c|}{R2R-CE Val-Unseen} & \multicolumn{4}{c}{RxR-CE Val-Unseen} \\
\cmidrule(lr){3-6} \cmidrule(lr){7-10}
 & & \textbf{NE}$\downarrow$ & \textbf{OS}$\uparrow$ & \textbf{SR}$\uparrow$ & \textbf{SPL}$\uparrow$ & \textbf{NE}$\downarrow$ & \textbf{SR}$\uparrow$ & \textbf{SPL}$\uparrow$ & \textbf{nDTW}$\uparrow$ \\
\midrule
4 & 3 & 4.92 & 62.1 & 56.4 & 51.2 & 5.30 & 55.2 & 45.3 & 59.5 \\
\rowcolor{LightGray}
8 & 3 & \textbf{4.57} & \textbf{64.9} & \textbf{60.3} & \textbf{55.1} & \textbf{4.90} & \textbf{60.3} & \textbf{49.8} & \textbf{62.1} \\
16 & 3 & 4.59 & 64.7 & 60.1 & 54.8 & 4.93 & 59.9 & 49.5 & 61.8 \\
\midrule
8 & 1 & 4.81 & 63.5 & 58.1 & 53.4 & 5.15 & 57.5 & 47.1 & 60.2 \\
8 & 5 & 4.60 & 64.8 & 60.0 & 54.9 & 4.92 & 60.1 & 49.6 & 62.0 \\
\bottomrule
\end{tabular}}
\end{table}

%% file: tables/tab_ablation_pgsa.tex
\begin{table}[t]
\centering
\footnotesize
\setlength{\tabcolsep}{4pt}
\caption{\textbf{Ablation study on the perception modules within the PGSA Auditor.} \textbf{Global}: CLIP~\cite{radford2021learning} Image-Text Similarity; \textbf{BBox}: GroundingDINO~\cite{liu2024grounding} object detection; \textbf{Mask}: SAM3~\cite{carion2025sam3segmentconcepts} precise segmentation.}
\label{tab:ablation_pgsa}
\resizebox{\linewidth}{!}{
\begin{tabular}{c|ccc|cccc|cccc}
\toprule
\multirow{2}{*}{Row} & \multicolumn{3}{c|}{Perception Modules} & \multicolumn{4}{c|}{R2R-CE Val-Unseen} & \multicolumn{4}{c}{RxR-CE Val-Unseen} \\
\cmidrule(lr){2-4} \cmidrule(lr){5-8} \cmidrule(lr){9-12}
 & \textbf{Global} & \textbf{BBox} & \textbf{Mask} & \textbf{NE}$\downarrow$ & \textbf{OS}$\uparrow$ & \textbf{SR}$\uparrow$ & \textbf{SPL}$\uparrow$ & \textbf{NE}$\downarrow$ & \textbf{SR}$\uparrow$ & \textbf{SPL}$\uparrow$ & \textbf{nDTW}$\uparrow$ \\
\midrule
1 & $\checkmark$ & & & 4.95 & 63.2 & 57.6 & 52.1 & 5.45 & 56.5 & 46.2 & 59.5 \\ 
2 & $\checkmark$ & $\checkmark$ & & 4.71 & 64.1 & 59.2 & 54.0 & 5.11 & 58.7 & 48.3 & 61.1 \\ 
\rowcolor{LightGray}
3 & $\checkmark$ & $\checkmark$ & $\checkmark$ & \textbf{4.57} & \textbf{64.9} & \textbf{60.3} & \textbf{55.1} & \textbf{4.90} & \textbf{60.3} & \textbf{49.8} & \textbf{62.1} \\ 
\bottomrule
\end{tabular}}
\end{table}

%% file: sections/Conclusion.tex
\section{Conclusion}
\label{sec:conclusion}

In this work, we introduce Step-Aware Contrastive Alignment (SACA), a novel reinforcement fine-tuning framework designed to overcome learning-signal collapse in sparse-reward Vision-Language Navigation. By leveraging a zero-shot Perception-Grounded Step-Aware (PGSA) auditor, SACA effectively extracts dense, step-level supervision for both continuous ranking scores and discrete structural boundaries, bypassing the need for costly task-specific Process Reward Models. The Scenario-Conditioned Group Construction mechanism seamlessly integrates Repair Resampling for near-misses and All-Failure Rescue for entirely failed batches. The robust SACA optimization objective fuses trajectory-level advantages and step-level constraints to rescue dense supervision from failed rollouts, preventing RL signal collapse. Extensive experiments on VLN-CE benchmarks demonstrate that SACA establishes a new state-of-the-art performance. 

%% file: sections/Appendix.tex
\begin{center}
    \author{}
    \institute{}
    \title{\inlineicon{pictures/career.png}{1.3em}{-0.5em}Let's Reward Step-by-Step: Step-Aware Contrastive Alignment for Vision-Language Navigation in Continuous Environments} 
    \titlerunning{\inlineicon{pictures/career.png}{1.3em}{-0.5em}Let's Reward Step-by-Step}
    \maketitle
    \vspace{-2em}
    {\Large \textit{Supplementary Material}}
    
\end{center}

\setcounter{figure}{0}
\setcounter{table}{0}
\setcounter{equation}{0}

\renewcommand{\thefigure}{S\arabic{figure}}
\renewcommand{\thetable}{S\arabic{table}}
\renewcommand{\theequation}{S\arabic{equation}}

\renewcommand{\theHsection}{appendix.\arabic{section}} 
\renewcommand{\theHfigure}{S\arabic{figure}}
\renewcommand{\theHtable}{S\arabic{table}}
\renewcommand{\theHequation}{S\arabic{equation}}

\noindent
\begin{tabular}{@{}p{0.03\textwidth} p{0.88\textwidth} @{}p{0.05\textwidth}@{}}
~\ref{app:statistics} & Statistics of Failed Episodes \dotfill & 20 \\[1em]

~\ref{app:more_details_SACA} & More Details about SACA Framework \dotfill & 21 \\[1em]
  & \hspace{1em} ~\ref{app:pgsa} \hspace{0.5em} Implementation Details of the PGSA Auditor \dotfill & 21 \\[0.8em]
  & \hspace{1em} ~\ref{app:saca_optimization} \hspace{0.5em} Execution Dynamics of the SACA Optimization \dotfill & 21 \\[0.8em]
  & \hspace{1em} ~\ref{app:step_level_reward} \hspace{0.5em} Anatomy of Step-Level Reward Allocation in SACA \dotfill & 22 \\[1em]

~\ref{app:hyperparams} & Implementation Details and Hyperparameters \dotfill & 23 \\[1em]

~\ref{app:visualize} & More Visualize Examples \dotfill & 26
\\[1em]

~\ref{app:prompt_design} & Prompt Design for the SACA Framework \dotfill & 26
\\[1em]

& \hspace{1em} ~\ref{app:instruction} \hspace{0.5em} Instruction Parsing Prompt for Landmark Extraction \dotfill & 26 \\[0.8em]
& \hspace{1em} ~\ref{app:action prediction} \hspace{0.5em} Action Prediction Prompt for SACA Training \dotfill & 27 \\[1em]

\end{tabular}
\vspace{2em} 

\section{Statistics of Failed Episodes}
\label{app:statistics}

\begin{wrapfigure}{r}{0.6\textwidth}
    \centering
    \includegraphics[width=0.6\textwidth]{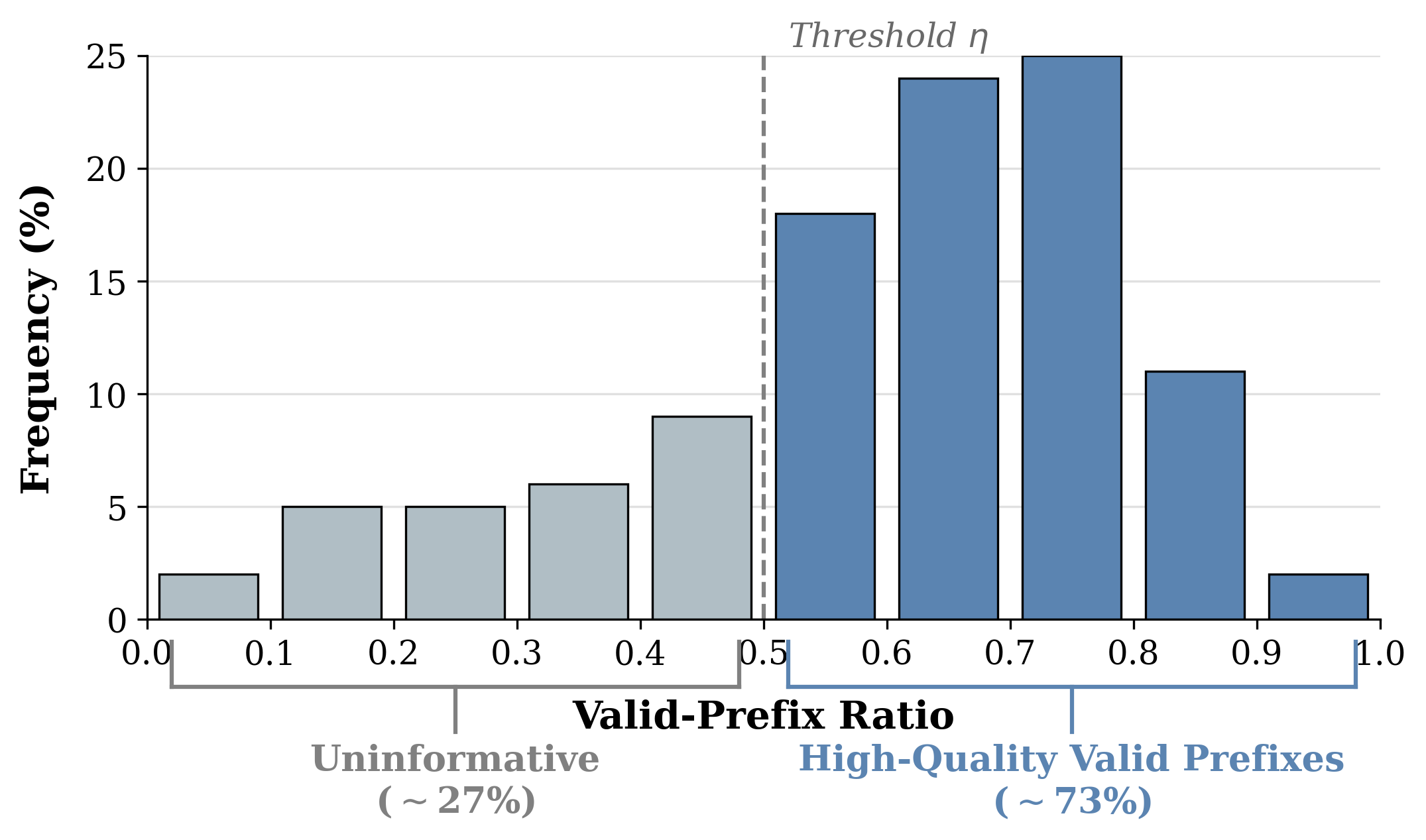}
    \vspace{-1em}
    \caption{\textbf{Distribution of valid-prefix ratios among failed navigation trajectories.} }
    \label{fig:distribution}
    \vspace{-2em}
\end{wrapfigure}
As shown in Fig.~\ref{fig:distribution}, our statistics indicates that a significant majority ($\sim 73\%$) of failed episodes actually contain a high-quality valid prefix before the agent deviates from the optimal path ($t_{div}/T > \eta$, with $\eta=0.5$). While standard RFT methods naively discards these near-miss failures entirely, SACA actively identifies and salvages these valuable prefixes to provide dense step-level supervision, thereby significantly mitigating data inefficiency.
\label{fig:prefix_ratio}

\section{More Details about Step-Aware Contrastive Alignment Framework}

We provide the granular implementation mechanics and algorithmic intricacies corresponding to Algorithm~\ref{alg:pgsa_auditor} and Algorithm~\ref{alg:saca_grpo}.
\label{app:more_details_SACA}

\subsection{Implementation Details of the PGSA Auditor} 
\label{app:pgsa}

Algorithm~\ref{alg:pgsa_auditor} formalizes the step-wise auditing process. Evaluating continuous visual streams against textual landmarks requires several adaptations:

\noindent \textbf{1. Prompting Strategy for Zero-Shot Grounding ($\mathcal{M}_{det}$):} 
To initialize the detection module  (GroundingDINO~\cite{liu2024grounding}), the raw intermediate landmark string $l_i$ is wrapped into a standardized prompt template: ``A photo of a [$l_i$].'' to align with the model's pre-training distribution. If GroundingDINO~\cite{liu2024grounding} returns multiple bounding boxes exceeding $\tau_{det}$, we solely retain the instance with the highest confidence score $c_{det}$ to compute the local similarity $s_{local}$, thereby suppressing background noise.

\noindent \textbf{2. Local Masking Operation ($o_t \odot M_{obj}$):} 
We utilize the pixel-level mask $M_{obj}$ generated by $\mathcal{M}_{seg}$ (SAM3~\cite{carion2025sam3segmentconcepts}). The background pixels outside $M_{obj}$ are strictly zeroed out (padded with black pixels) before being encoded by the CLIP visual encoder $\mathcal{M}_{vlm}$. This strict isolation ensures that $s_{local}$ genuinely reflects the semantic alignment of the target object alone.

\noindent \textbf{3. Landmark Transition Mechanism:}
Algorithm~\ref{alg:pgsa_auditor} evaluates a single active landmark $l_i$. In our full pipeline, the instruction comprises a sequence of landmarks $L = \{l_1, \dots, l_m\}$, generated by frozen Qwen3-0.6B~\cite{yang2025qwen3}. The auditor transitions from $l_i$ to $l_{i+1}$ when a robust temporal-spatial constraint is met: the step-wise Soft Score $S_t$ must exceed the hard threshold $\tau_h$ for $c=2$ consecutive steps. This temporal smoothing prevents premature landmark transitions caused by transient visual glitches or motion blur during navigation.

\subsection{Execution Dynamics of the SACA Optimization}
\label{app:saca_optimization}

Algorithm~\ref{alg:saca_grpo} details the routing and policy update mechanism. To deploy this efficiently on MLLMs, several systemic optimizations and definitions are applied:

\noindent \textbf{1. Definition of Outcome Reward ($r^{out}$):} 
The binary outcome reward $r^{out}(\mathcal{T}_i)$ is strictly determined by the environment's topological graph. It evaluates to $1$ only if the agent asserts the \texttt{STOP} action and its final spatial coordinates are within a constant value \textit{e.g.} $3.0$ meters of the destination.

\noindent \textbf{2. GRPO KL-Divergence Penalty:} 
In both the Mixed Group (Eq.~\ref{eq:l_mixed}) and All-Failure Rescue (Eq.~\ref{eq:l_fail}) scenarios, the $\mathcal{L}_{GRPO}$ function incorporates an implicit Kullback-Leibler (KL) divergence penalty to prevent policy collapse. Specifically, the reward advantages $\hat{A}_i$ are dynamically adjusted using $\hat{A}_i - \beta_{KL} \text{KL}(\pi_{\theta} \| \pi_{ref})$, where the reference policy $\pi_{ref}$ is initialized from the SFT checkpoint and frozen. The KL penalty ensures the policy remains exploratory and prevents deterministic mode collapse during intensive Repair Resampling.

\noindent \textbf{3. Tie-Breaking for the Pseudo-Anchor:} 
In Scenario B, the framework selects the Pseudo-Anchor $\mathcal{T}_{anch}$ based on the maximum process score $R_{proc}(\mathcal{T})$. In the rare event of a tie where multiple trajectories share the exact same $R_{proc}$ score, we employ the maximum trajectory length $T_i$ as a secondary tie-breaking criterion, favoring the trajectory that explored further before failing.

\subsection{Anatomy of Step-Level Reward Allocation in SACA}
\label{app:step_level_reward}

Table~\ref{tab:reward_summary} summarizes the temporal allocation of reward signals and their corresponding optimization objectives across different phases of a repaired trajectory.

\begin{table}[h]
    \centering
    \vspace{-2em}
    \caption{\textbf{Summary of Step-Level Reward and Objective Allocation in SACA.}}
    \label{tab:reward_summary}
    \resizebox{\textwidth}{!}{
    \begin{tabular}{lcccc}
        \toprule
        \textbf{Trajectory Phase} & \textbf{Temporal Condition} & \textbf{Hard Mask} & \textbf{Dominant Reward Signal ($\hat{A}_t$)} & \textbf{Optimization Objective} \\
        \midrule
        \textbf{Valid Prefix} & $t < t_{div}$ & $M_t = 1$ & \cellcolor{red!10} Dense Positive (Continuous Baseline + Peaks) & $\mathcal{L}_{align}$ (Behavior Cloning) \\
        \textbf{Divergence Point} & $t = t_{div}$ & $M_t = 0$ & \cellcolor{blue!10} Sharp Negative (Explicit Penalty) & $\mathcal{L}_{corr}$ (Contrastive Correction) \\
        \textbf{Resampled Suffix} & $t > t_{div}$ & $M_t = 1$ & \cellcolor{red!10} Dense Positive (Continuous Baseline + Peaks) & $\mathcal{L}_{repair}$ (Suffix Cloning) \\
        \bottomrule
    \end{tabular}
    }
    \vspace{-2em}
\end{table}

\noindent \textbf{1. Dense Positive Alignment ($t \neq t_{div}$).} For all steps within the valid prefix and the successfully resampled suffix, the agent is considered to be on the optimal path ($M_t = 1$). During these phases, the optimization functions effectively treat the executed actions as pseudo-correct demonstrations. The underlying reward signal is driven by the Composite Soft Score $S_t$, which provides a continuous positive advantage. This ensures the policy receives steady contextual guidance (via global semantics) interspersed with strong positive reinforcement peaks (via local spatial grounding) upon encountering landmarks.

\noindent \textbf{2. Explicit Contrastive Penalty ($t = t_{div}$).} The critical inflection point occurs exactly at the divergence step. When $S_t \le \tau_h$, the mask flips ($M_t = 0$), signaling a fatal deviation. SACA isolates this specific step. The Contrastive Correction objective injects a sharp, explicit negative penalty strictly at $t_{div}$, pulling the policy towards the correct teacher action $a^+$ while aggressively pushing it away from the specific erroneous action $a^-$ that caused the deviation.

\begin{wrapfigure}{r}{0.5\textwidth}
    \vspace{-2em}
    \centering
    \includegraphics[width=0.5\textwidth]{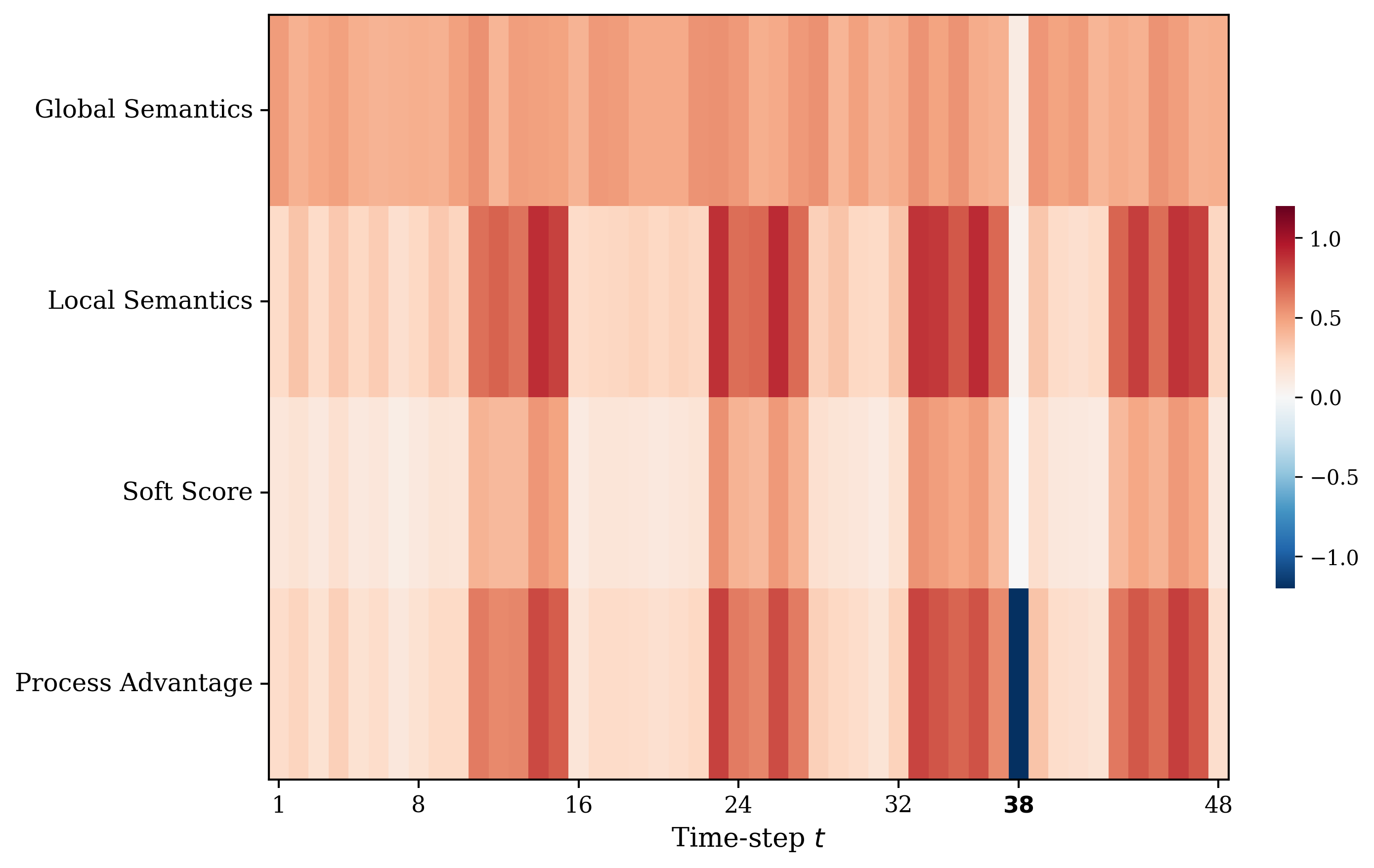}
    \vspace{-1em}
    \caption{\textbf{Heatmap visualization.} We visualize the step-wise evolution of semantic alignment and advantage assignment over a trajectory consist with $48$ steps.}
    \label{fig:signal_heatmap}
    \vspace{-2em}
\end{wrapfigure}
As shown in Fig.~\ref{fig:signal_heatmap}, SACA assigns heterogeneous signals based on the structural phase of the trajectory, seamlessly transitioning from dense positive alignment to explicit contrastive penalty. During the \texttt{Valid Prefix} ($t < 38$) and the \texttt{Resampled Suffix} ($t > 38$), the fusion of continuous Global Semantics and sharp Local Semantics provides a dense, positive Process Advantage (warm colors). At the \texttt{Divergence Point} ($t_{div} = 38$), a sudden drop in semantic matching causes the Composite Soft Score to plummet, which in turn triggers the structural hard mask. This explicitly applies a Contrastive Correction penalty (deep blue), heavily discouraging the precise erroneous action before the trajectory is successfully resampled. By structuring the step-level rewards in this heterogeneous manner, SACA elegantly resolves the credit assignment problem in long-horizon navigation, ensuring that the agent is neither penalized for correct early decisions nor left unguided after a mid-trajectory failure.

\input{algorithms/alg_pgsa_auditor}
\input{algorithms/alg_saca_grpo}

\section{Implementation Details and Hyperparameters}
\label{app:hyperparams}

To ensure the full reproducibility of our experiments, Table~\ref{tab:hyperparameters} meticulously details the comprehensive hyperparameter configurations utilized across both SFT and SACA stages.

\begin{itemize}[leftmargin=*, label=\textbullet]
{
\item \textbf{General Infrastructure.} All training phases are distributed across 8 GPUs utilizing the Distributed Data Parallel (DDP) strategy. To accommodate the extensive context required for continuous visual observations and long-horizon instructions, we set the maximum sequence length to 4096. Furthermore, we leverage bfloat16 mixed precision and Flash Attention to significantly optimize memory consumption and accelerate training throughput.

\item \textbf{SFT Stage.} During the preliminary SFT phase, the policy is trained for 2 epochs using the AdamW optimizer with $\beta$ values of $[0.9, 0.999]$ and a weight decay of 0.01. We employ a constant learning rate of $1 \times 10^{-5}$ and maintain a total global batch size of 16.

\item \textbf{Reinforcement Fine-Tuning (SACA) Stage.} In the subsequent reinforcement learning phase, the learning rate is strictly reduced to $1 \times 10^{-6}$ to ensure stable policy updates. For the GRPO rollout, we sample $K=8$ candidate trajectories per instruction. To salvage near-miss failures, the valid-prefix ratio threshold ($\eta$) is set to 0.5, and the policy is permitted a maximum of $N_{rep}=3$ suffix resampling attempts. For the PGSA Auditor, we rigorously calibrate the sensitivity parameters to balance exploration and precise grounding. Specifically, we apply a soft threshold $\tau_s=0.2$ for step-score noise filtering, a hard threshold $\tau_h=0.25$ for structural divergence masking, and a detection threshold $\tau_{det}=0.3$ to filter GroundingDINO~\cite{liu2024grounding} bounding boxes. The step-wise soft score components are balanced by weights $w_1=0.3$, $w_2=0.3$, and $w_3=0.4$. Finally, to regularize the advantage estimation and prevent over-penalization, we implement robust scaling mechanisms featuring a margin threshold $\delta=0.1$, a temperature shrinkage factor $\kappa=0.5$, and a negative attenuation scale $s=0.5$.
} 
\end{itemize}

\input{tables/tab_hyperparameters}

\section{More Visualize Examples}
\begin{figure}[b]
    \centering
    \includegraphics[width=\linewidth]{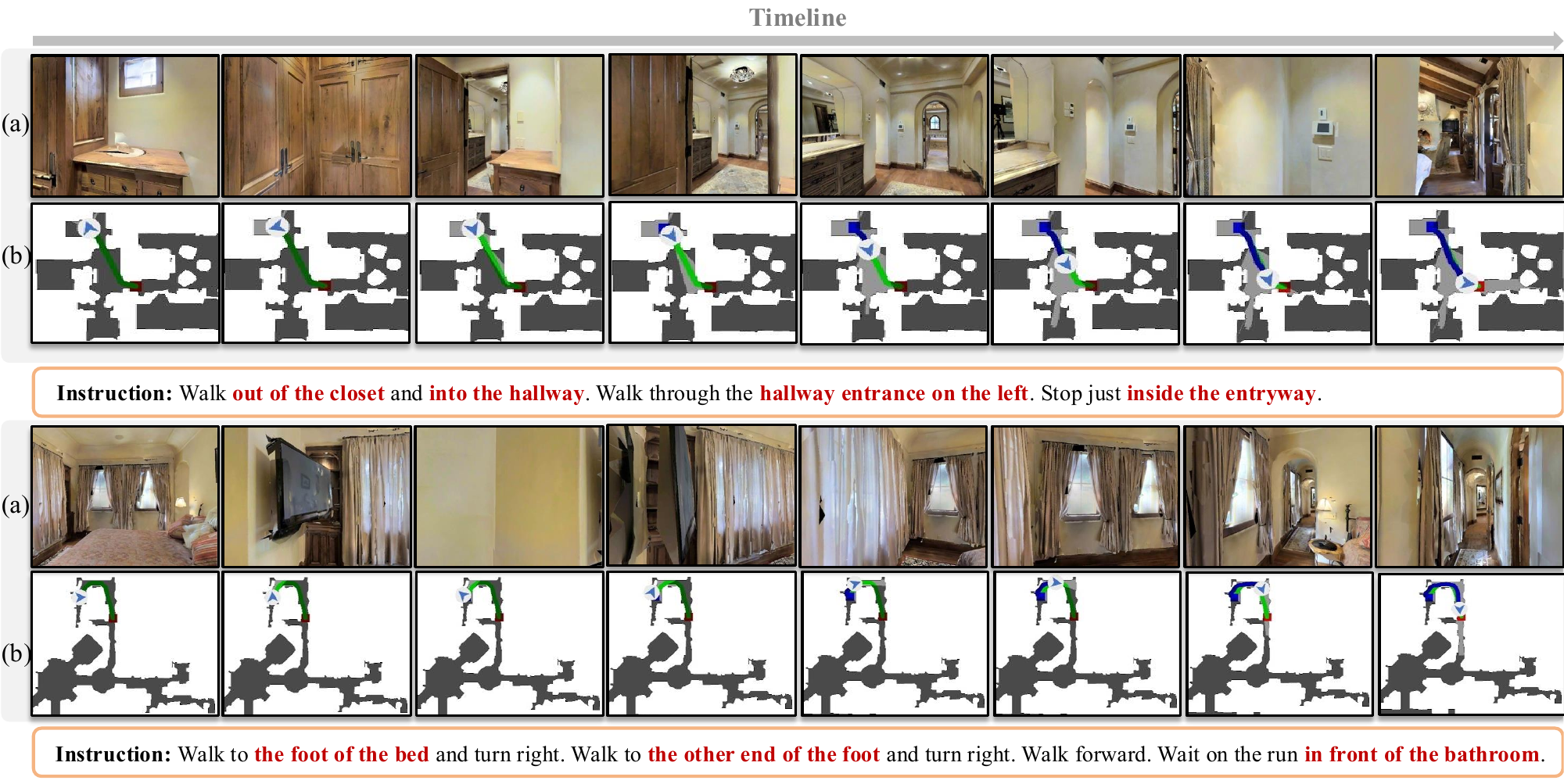}
    \caption{\textbf{Qualitative visualization of navigation episodes.} \textbf{(a)} Continuous egocentric visual observations. \textbf{(b)} The corresponding top-down trajectory maps. Key textual landmarks parsed from the instructions are highlighted in red, guiding the step-aware progress tracking.}
\label{fig:more_visualize}
\end{figure}

\begin{figure}[t]
    \centering
    \includegraphics[width=\linewidth]{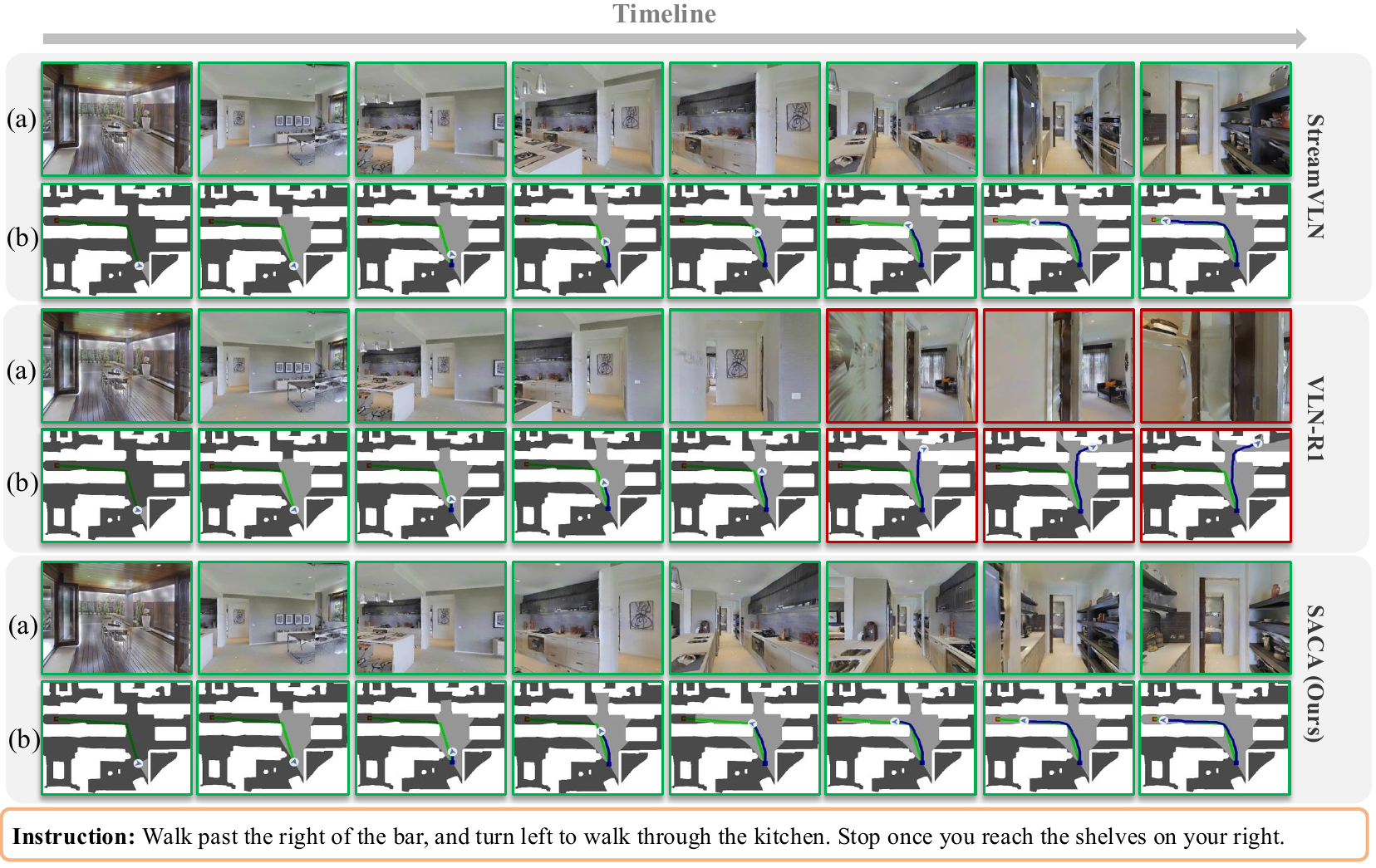}
    \caption{\textbf{More qualitative comparison.} \textbf{(a)} Egocentric observations. \textbf{(b)} Top-down maps. Green and red frames denote correct trajectory and failure trajectory, respectively.}
    \label{fig:conparison_2}
\end{figure}

We provide more qualitative visualization (Fig.~\ref{fig:more_visualize}) and more qualitative comparison results with other methods \textit{e.g.} StreamVLN~\cite{wei2026streamvln}, VLN-R1~\cite{qi2025vln} (Fig.~\ref{fig:conparison_2}).
\label{app:visualize}

\section{Prompt Design for the SACA Framework}
\label{app:prompt_design}

In this section, we provide the detailed prompt templates utilized throughout our SACA framework. Specifically, we first introduce the instruction parsing prompt used by the frozen tiny LLM to extract intermediate landmarks, followed by the action prediction prompt employed by the policy MLLM during the RFT phase.

\definecolor{promptbg}{RGB}{248, 248, 248}
\definecolor{promptframe}{RGB}{102, 102, 102}

\newtcolorbox{promptbox}[1]{
    colback=promptbg,
    colframe=promptframe,
    fonttitle=\bfseries,
    title=#1,
    boxrule=1.5pt,
    arc=4pt,
    left=8pt, right=8pt, top=8pt, bottom=8pt,
    breakable 
}

\subsection{Instruction Parsing Prompt for Landmark Extraction}
\label{app:instruction}
To enable the PGSA auditor to track navigation progress, we first parse the high-level linguistic instruction into a sequence of intermediate landmarks $L = \{l_1, \dots, l_m\}$. As detailed in Box 1, we deploy a frozen tiny LLM (\textit{e.g.}, Qwen3-0.6B~\cite{yang2025qwen3}) with a structured zero-shot prompt to extract these landmarks sequentially.

\begin{promptbox}{Box 1. Instruction Parsing Prompt for Tiny LLM}
\textbf{System}: You are a specialized linguistic parser for an embodied navigation agent. Your task is to extract a sequential list of intermediate visual landmarks from a given navigation instruction. These landmarks will be used by a downstream visual perception module to track the agent's progress.

\vspace{0.5em}
\noindent\textbf{User}: 
 \texttt{<Instruction>}

\vspace{0.5em}
\noindent\textbf{Output Format}:
You must extract the physical objects or locations mentioned in the instruction in the exact order they are intended to be encountered. Output strictly a valid JSON list of strings. Do not include any other explanations.

\vspace{0.5em}
\noindent\textbf{Example Input}: Walk past the dining room table and chairs and turn right. Walk past the bar and stop by the bar stools.

\vspace{0.5em}
\noindent\textbf{Example Output}: \texttt{[table, chairs, bar, bar stools]}
\end{promptbox}

\vspace{1em}

\subsection{Action Prediction Prompt for SACA Training}
\label{app:action prediction}
During the SACA training phase, we aim to enable the policy MLLM to effectively align its visual perception with the linguistic instruction. To this end, we design a concise, structured prompt that incorporates both current and historical visual observations, explicitly defines the discrete action options, and directly outputs the action sequence. The complete user input format is presented in Box 2. 

\begin{promptbox}{Box 2. System Prompt and Input Format for SACA Training}
\textbf{System}: You are an intelligent embodied agent navigating in a continuous 3D indoor environment. Your objective is to reach a specific destination based on a natural language instruction. At each step, you will be provided with the instruction and your current egocentric visual observation. Your task is to analyze the visual scene, align it with the given instruction, reason about your current progress, and predict the next optimal action.

\vspace{0.5em}
\noindent\textbf{User}: 
\texttt{<Image>, <Instruction>} 

\vspace{0.5em}
\noindent\textbf{Output Format}:
Devise an action sequence to follow the instruction using the four actions: TURN LEFT ($\leftarrow$) or TURN RIGHT ($\rightarrow$) by 15 degrees, MOVE FORWARD ($\uparrow$) by 25 centimeters, or \texttt{STOP}.

\vspace{0.5em}
\noindent\textbf{Example}: 
\texttt{<action>}$\rightarrow\rightarrow\rightarrow\uparrow$\texttt{</action>}

\end{promptbox}

%% file: algorithms/alg_pgsa_auditor.tex
\begin{algorithm}[htbp]
\caption{Perception-Grounded Step-Aware (PGSA) Auditing Process}
\label{alg:pgsa_auditor}
\begin{algorithmic} 
\Require Trajectory $\mathcal{T} = \{o_1, \dots, o_T\}$; Current target landmark ($l_{\text{i}}$)
\Require GroundingDINO ($\mathcal{M}_{det}$), SAM2 ($\mathcal{M}_{seg}$), CLIP ($\mathcal{M}_{vlm}$)
\Require Hyperparameters: Detection threshold ($\tau_{det}$); Soft threshold ($\tau_s$); Hard threshold ($\tau_h$); Weights $w_1, w_2, w_3$
\State Initialize process score $R_{proc} \leftarrow 0$
\State Initialize divergence point $t_{div} \leftarrow T$, divergence flag $F_{div} \leftarrow \text{False}$
\State Initialize Hard Mask sequence $M \leftarrow \{1, 1, \dots, 1\}$ of length $T$

\For{$t = 1$ to $T$}
    
    \begin{mybox}{Phase 1: Semantic Grounding \& Global Context}
    $B, c_{det} \leftarrow \mathcal{M}_{det}(o_t, l_{\text{i}})$ \Comment{Detect bounding box and confidence}
    $s_{global} \leftarrow \mathcal{M}_{vlm}(o_t, l_{\text{i}})$ \Comment{Compute global image-text similarity}
    \end{mybox}
    
    \begin{mybox}{Phase 2 : Spatial Refinement \& Local Alignment}
    \If{$c_{det} \geq \tau_{det}$}
        \State $M_{obj}, c_{iou} \leftarrow \mathcal{M}_{seg}(o_t, B)$ \Comment{Extract precise object mask}
        \State $s_{local} \leftarrow \mathcal{M}_{vlm}(o_t \odot M_{obj}, l_{\text{i}})$ \Comment{Compute local masked similarity}
    \Else
        \State $s_{local} \leftarrow 0$, $c_{iou} \leftarrow 0$ \Comment{Fallback if object is totally invisible}
    \EndIf
    \end{mybox}
    
    \begin{mybox}{Phase 3: Comprehensive Soft Score Calculation}
    $S_t \leftarrow w_1 s_{global} + \mathbb{1}_{\{c_{det} \geq \tau_{det}\}} \big( w_2 c_{det} + w_3 c_{iou} s_{local} \big)$ \Comment{(Eq.~\ref{eq:soft_score})}
    \end{mybox}
    
    \begin{mybox}{Phase 4: Accumulation \& Masking}
    \If{$S_t \geq \tau_s$}
        \State $R_{proc} \leftarrow R_{proc} + (S_t - \tau_s)$ \Comment{Accumulate continuous process reward}
    \EndIf
    
    \If{$S_t < \tau_h$ \textbf{and not} $F_{div}$}
        \State $t_{div} \leftarrow t$ \Comment{Locate the precise structural Divergence Point (Eq.~\ref{eq:t_div})}
        \State $F_{div} \leftarrow \text{True}$
    \EndIf
    
    \If{$F_{div}$}
        \State $M_t \leftarrow 0$ \Comment{Apply Hard Mask}
    \EndIf
    \end{mybox}

\EndFor

\State $R_{proc} \leftarrow R_{proc} / T$ \Comment{Normalize process score over trajectory length (Eq.~\ref{eq:r_proc})}
\State \Return $R_{proc}$, $t_{div}$, $M$
\end{algorithmic}
\end{algorithm}

%% file: algorithms/alg_saca_grpo.tex
\begin{algorithm}[htbp]
\caption{Step-Aware Contrastive Alignment (SACA) Framework}
\label{alg:saca_grpo}
\begin{algorithmic} 
\Require Policy $\pi_\theta$; Reference policy $\pi_{\text{ref}}$; PGSA Auditor $\mathcal{A}_{\text{PGSA}}$ (Algorithm~\ref{alg:pgsa_auditor})
\Require Hyperparameters: Group size $K$; Sub-group size $m$; Valid ratio $\eta$; Margin $\delta$; Temp $\kappa$; Scale $s$; Max repairs $N_{rep}$

\While{training is not converged}
    \State Sample instruction $I$ and extract intermediate landmarks $L = \{l_1, \dots, l_m\}$
    \State Sample $K$ candidate trajectories $G = \{\mathcal{T}_1, \dots, \mathcal{T}_K\} \sim \pi_{\theta}(\cdot \mid I)$
    
    \begin{mybox}{Phase 1: Perception-Grounded Auditing}
    \textbf{for all} $\mathcal{T}_i \in G$ \textbf{do} \\
    \quad Compute soft score $R_{proc}(\mathcal{T}_i)$ (Eq.~\ref{eq:r_proc}), divergence point $t_{div}^{(i)}$ (Eq.~\ref{eq:t_div}), and mask $M^{(i)}$  via $\mathcal{A}_{\text{PGSA}}(\mathcal{T}_i, L)$ \\
    \quad Evaluate binary outcome reward $r^{out}(\mathcal{T}_i) \in \{0, 1\}$
    \end{mybox}
     
        \begin{mybox}{Phase 2a: Scenario A - Mixed Group}
        Compute outcome advantages $\hat{A}^{out}_i$ normalized over the full group $G$ (Eq.~\ref{eq:a_out}) \\
        Initialize repaired group $G_{rep} \leftarrow \emptyset$ \\
        \textbf{for all} $\mathcal{T}_j \in G$ s.t. $r^{out}(\mathcal{T}_j) = 0$ \textbf{and} $t_{div}^{(j)} / T > \eta$ \textbf{do} \\
        \quad Sample suffixes $\{\mathcal{T}_{suff}^{(n)}\}_{n=1}^{N_{rep}} \sim \pi_{\theta}(\cdot \mid \mathcal{T}_{j, 1:t_{div}^{(j)}})$ \\
        \quad \textbf{if} $\exists n \in \{1, \dots, N_{rep}\}$ s.t. $r^{out}(\mathcal{T}_{suff}^{(n)}) = 1$ \textbf{then} \\
        \qquad Construct $\mathcal{T}^{\text{rep}}_j \leftarrow \mathcal{T}_{j, 1:t_{div}^{(j)}} \oplus \mathcal{T}_{suff}^{(n)}$ \hfill $\triangleright$ $\oplus$ denotes concatenation \\
        \qquad $G_{rep} \leftarrow G_{rep} \cup \{\mathcal{T}^{\text{rep}}_j\}$ \\
        Compute total loss $\mathcal{L}_{\text{total}} \leftarrow \mathcal{L}_{mixed}(\hat{A}^{out}, G_{rep})$ (Eq.~\ref{eq:l_mixed})
        \end{mybox}
        
        \begin{mybox}{Phase 2b: Scenario B - All-Failure Rescue}
        Select Pseudo-Anchor: $\mathcal{T}_{anch} \leftarrow \arg\max_{\mathcal{T} \in G} R_{proc}(\mathcal{T})$ (Eq.~\ref{eq:anchor}) \\
        Form negative subset $G_{neg} \subset G \setminus \{\mathcal{T}_{anch}\}$ comprising top-$m$ trajectories ranked by $\text{score}_{neg}$ (Eq.~\ref{eq:score_neg}) \\
        Construct Reflection Sub-group: $G_{sub} \leftarrow \{\mathcal{T}_{anch}\} \cup G_{neg}$ \\
        Compute raw advantages $\hat{A}^{raw}_i$ normalized over $G_{sub}$ (Eq.~\ref{eq:a_raw}) \\[0.5em]
        \textbf{Margin-Based Rescue:} $\hat{A}_i' \leftarrow \hat{A}_i^{raw} \cdot (\kappa \cdot \mathbb{I}_{[\Delta < \delta]} + \mathbb{I}_{[\Delta \ge \delta]})$ (Eqs.~\ref{eq:margin}, \ref{eq:a_prime}) \\
        \textbf{Negative-Only Scaling:} $\hat{A}_i \leftarrow \hat{A}_i' \cdot (s \cdot \mathbb{I}_{[\hat{A}_i' < 0]} + \mathbb{I}_{[\hat{A}_i' \ge 0]})$ (Eq.~\ref{eq:a_final}) \\[0.5em]
        Compute $\mathcal{L}_{align}$ and $\mathcal{L}_{correct}$ for $\mathcal{T}_{anch}$ leveraging divergence point $t_{div}^{(anch)}$ (Eqs.~\ref{eq:l_align}, \ref{eq:l_corr}) \\
        Compute total loss $\mathcal{L}_{\text{total}} \leftarrow \mathcal{L}_{fail}(\hat{A}_i, \mathcal{L}_{align}, \mathcal{L}_{correct})$ (Eq.~\ref{eq:l_fail})
        \end{mybox}
    
    \begin{mybox}{Phase 3: Policy Update}
    Update policy parameters: $\theta \leftarrow \theta - \alpha \nabla_\theta \mathcal{L}_{\text{total}}$
    \end{mybox}
    
\EndWhile
\end{algorithmic}
\end{algorithm}

%% file: tables/tab_hyperparameters.tex
\begin{table}[htbp]
    \centering
    \caption{\textbf{Comprehensive training configurations for SFT and SACA stages.}}
    \label{tab:hyperparameters}
    \resizebox{\textwidth}{!}{
    \begin{tabular}{lc|lc}
        \toprule
        \multicolumn{2}{c|}{\textbf{General \& SFT Configuration}} & \multicolumn{2}{c}{\textbf{SACA Hyperparameters}} \\
        \midrule
        \textbf{Hyperparameter} & \textbf{Value} & \textbf{Hyperparameter} & \textbf{Value} \\
        \midrule
        \multicolumn{2}{l|}{\textbf{General Infrastructure}} & \multicolumn{2}{l}{\textbf{Scenario \& Grouping}} \\
        Policy Model & Video-LLaVA-8B & Group size ($K$) & 8 \\
        Max Sequence Length & 4096 & Valid-prefix ratio ($\eta$) & 0.5 \\
        Number of GPUs & 8 & Max resampling attempts ($N_{rep}$) & 3 \\
        Parallel Strategy & DDP & \multicolumn{2}{l}{\textbf{PGSA Auditor Settings}} \\
        Training Precision & bfloat16 & Soft threshold ($\tau_s$) & 0.2 \\
        Flash Attention & \checkmark & Hard threshold ($\tau_h$) & 0.25 \\
        \multicolumn{2}{l|}{\textbf{SFT Stage Configuration}} & Detection threshold ($\tau_{det}$) & 0.3 \\
        Optimizer & AdamW & Score weights ($w_1, w_2, w_3$) & 0.3, 0.3, 0.4 \\
        Betas & [0.9, 0.999] & \multicolumn{2}{l}{\textbf{Robustness \& RL Settings}} \\
        Weight Decay & 0.01 & Margin threshold ($\delta$) & 0.1 \\
        Learning Rate & $1 \times 10^{-5}$ & Temperature shrinkage ($\kappa$) & 0.5 \\
        Batch Size (Total) & 16 & Negative attenuation ($s$) & 0.5 \\
        Epochs & 2 & Learning Rate & $1 \times 10^{-6}$ \\
        Warmup Ratio & 0.03 & \multicolumn{2}{l}{\textbf{Objective Weights}} \\
        LR Scheduler & Cosine & Hard-Negative weight ($\lambda$) & 0.5 \\
        Max Grad Norm & 1.0 & Repair loss weight ($\lambda_{rep}$) & 1.0 \\
        Gradient Accumulation & 2 & Align loss weight ($\lambda_1$) & 1.0 \\
        Drop Path Rate & 0.1 & Contrastive loss weight ($\lambda_2$) & 1.0 \\
        \bottomrule
    \end{tabular}
    }
\end{table}